\documentclass[twoside]{article}

\usepackage[preprint]{aistats2027}
\usepackage[round]{natbib}
\renewcommand{\bibname}{References}
\renewcommand{\bibsection}{\subsubsection*{\bibname}}

\usepackage{amsmath,amsfonts,bm}

\def\eqref#1{equation~\ref{#1}}

\def\1{\bm{1}}

\DeclareMathAlphabet{\mathsfit}{\encodingdefault}{\sfdefault}{m}{sl}
\SetMathAlphabet{\mathsfit}{bold}{\encodingdefault}{\sfdefault}{bx}{n}

\newcommand{\E}{\mathbb{E}}

\newcommand{\Var}{\mathrm{Var}}

\newcommand{\Cov}{\mathrm{Cov}}

\DeclareMathOperator{\Tr}{Tr}

\usepackage{hyperref}
\usepackage{url}

\usepackage{amssymb}
\usepackage[utf8]{inputenc} 
\usepackage[T1]{fontenc}    
\usepackage{hyperref}       
\usepackage{url}            
\usepackage{booktabs}       
\usepackage{amsfonts}       
\usepackage{nicefrac}       
\usepackage{microtype}      
\usepackage{xcolor}         

\usepackage{wrapfig}

\usepackage{graphicx}

\usepackage{array}
\usepackage{multirow}
\usepackage{cancel}
\usepackage{mathtools}
\usepackage{amsthm}
\usepackage{subcaption}

\usepackage{algorithm,algpseudocode}

\theoremstyle{plain}
\newtheorem{lemma}{Lemma}

\newtheorem{corollary}{Corollary}
\theoremstyle{remark}

\newcommand{\W}{{\Delta W}}
\newcommand{\out}{Y}
\newcommand{\nn}{f}
\newcommand{\lhat}{\hat\lambda}
\newcommand{\that}{\hat\theta}

\begin{document}

%

%

\twocolumn[

\aistatstitle{Prior Predictive Matching for Bayesian LoRA}

\aistatsauthor{ Tomasz Kuśmierczyk \And Eliezer de Souza da Silva \And  Marcin Sendera }

\aistatsaddress{ \And  University of Coimbra, \\ CISUC/LASI, DEI \And Jagiellonian University \\ NASK -- National Research Institute} ]

\begin{abstract}
Bayesian variants of Low-Rank Adaptation (LoRA) provide principled uncertainty estimation when fine-tuning large foundation models,
but require specifying the prior over the low-rank weight updates.
We address this hyperparameter selection problem by introducing a closed-form prior-predictive estimator that operates before fine-tuning.

We analyze a second-order expansion of the network and quantify how the low-rank prior drives function-space deviations from the pretrained outputs.
We also show that these prior-predictive quantities depend on the prior only through the product of the adaptation rank and the fourth power of the prior scale. 
Because the training targets share a single draw of the latent weight update, we can leverage the covariance between example residuals to derive a moment-matching estimator.
Furthermore, our theoretical analysis establishes that the optimal prior precision scales as $r^{-1/2}$ with the rank $r$, allowing a single estimate to generalize across all ranks.

Controlled experiments across regression and classification tasks validate our approximation and show that our estimator recovers the true prior scale.
On RoBERTa-large, our approach improves predictive negative log-likelihood (NLL) and calibration over both MAP and marginal-likelihood selection without requiring validation data, while achieving competitive out-of-distribution detection.
\end{abstract}
\section{Introduction}
\label{sec:intro}

Fine-tuning large pretrained models with Low-Rank Adaptation (LoRA) \citep{hu2022lora} restricts parameter updates to a low-dimensional subspace. For a selected weight matrix $W_0$, LoRA replaces $W_0$ by $W_0+\W$, where $\W$ is constrained to have a low-rank factorization. Bayesian extensions of LoRA 
(e.g., \citealp{yang2024laplacelora})
place a prior over this low-rank update, typically through Gaussian priors on its factors, and use the resulting posterior to quantify uncertainty. This introduces two modeling choices: the rank of the update and the scale of its prior, $\sigma$ (or equivalently its precision $\lambda=\sigma^{-2}$). Together, these quantities determine the magnitude and complexity of parameter changes regarded as plausible before observing the fine-tuning data and can therefore have a substantial effect on the resulting posterior.

In existing approaches, these hyperparameters are typically selected either by validation search over the trained models or by maximizing an approximation to the marginal likelihood. 
Both strategies require fitting the model before the prior hyperparameters can be assessed. This creates a mismatch between the role of the prior and the way it is specified: a quantity intended to encode plausible adaptations \emph{before} fine-tuning is chosen using the outcome of fine-tuning itself. 

For Bayesian matrix factorization (BMF), this mismatch is avoided by prior-predictive matching, which sets the hyperparameters 
by matching moments of the prior predictive distribution to the data \citep{dasilva2023prior}.
LoRA decomposes the update as $\Delta W = UV$ and can therefore be viewed as a matrix factorization in parameter space. However, extending prior-predictive matching to LoRA is not immediate. In BMF, the observations are entries of the random matrix generated by the latent factors. In Bayesian LoRA, the observations are input-label pairs $(x,y)$, and the latent update affects them only through the nonlinear network. The resulting prior-predictive distribution exhibits entirely different characteristics. Consequently, Bayesian LoRA requires a fundamentally different treatment from standard BMF.


In this paper, we study the prior predictive distribution of Bayesian LoRA and analyze its moments. Our estimators allow the prior hyperparameters to be selected in closed form. 
They need only the pretrained model and the fine-tuning data, and require neither LoRA optimization nor validation data (Table~\ref{tab:needs} compares the approaches).
We effectively turn the prior specification from a model-selection problem after fine-tuning into a statistical estimation problem before fine-tuning.

Our contributions are as follows:
\begin{enumerate}\setlength{\itemsep}{0pt}
    \item We develop a prior-predictive analysis of Bayesian LoRA that explicitly relates the LoRA prior scale and rank to moments of the induced changes in network outputs.
    \item We derive a closed-form prior-predictive estimator for the Bayesian LoRA prior scale that can be computed before fine-tuning, does not require validation over trained models or post-hoc evidence optimization, and separates shared parameter-induced variation from independent observation noise.
    \item We evaluate the resulting calibration procedure across controlled and pretrained-model experiments, testing whether prior-predictive calibration recovers appropriate prior scales and whether these scales provide competitive uncertainty quantification and predictive performance after Bayesian LoRA fine-tuning.
\end{enumerate}
The code is available \href{https://github.com/tkusmierczyk/prior-predictive-lora}{here}. The appendices contain the proofs, additional estimators, the details of the experiments, and a
further study of fine-tuning under the estimated prior (Appendix~\ref{app:train-lambda}).

\section{Background}
\label{sec:setting}
\paragraph{Generative model.} 
We consider a pretrained neural network $\nn(x; \{W\})$ with frozen weights in which selected linear layers are adapted (fine-tuned) as follows:
$W_0 \to W_0 + \W$. The update factorizes as $\W = UV$ with $U \in \mathbb{R}^{N_o\times K}$ and
$V \in \mathbb{R}^{K\times N_i}$, where $K$ is the LoRA rank and $N_i$, $N_o$ are the layer input and output dimensions. The Bayesian model places the
prior on the factors as: 
\begin{equation}
U_{ik} \sim \mathcal N\big(\mu_U, \sigma_U^2\big), \qquad
V_{kj} \sim \mathcal N\big(\mu_V, \sigma_V^2\big),
\label{eq:factorprior}
\end{equation}
where $\mu_U$ and $\sigma_U$ denote the mean and scale (standard deviation) shared by all entries of
$U$, and $\mu_V$ and $\sigma_V$ the same for $V$. 
Throughout most of the paper 
we assume the standard (e.g., \citealp{yang2024laplacelora}) zero means and one shared
scale: $\mu_U = \mu_V = 0$, $\sigma_U = \sigma_V = \sigma$. 
The prior $p(\W \mid \sigma, K)$ is then \emph{induced} from \eqref{eq:factorprior} through $\Delta W = U V$. 



Throughout the rest of the paper,
for brevity,  
we consider the frozen weights to be part of the network function itself and focus only on one-layer updates: 
$\nn(x; W_0 + \Delta W) \equiv \nn(x; \Delta W)$.
We explain the generalization to multiple modules in
Section~\ref{sec:implementation}.
\\
For random $\Delta W$, the output $\out(x) = \nn(x; \W)$ is also a random variable conditioned on the input $x$. 
Therefore, the full generative model for 
fine-tuning is:
{\small
\begin{equation}
\begin{aligned}
&\W^\ast \sim p(\W \mid \sigma,K)\; \text{once,}\;
y_n \sim p\big(y\mid\nn(x_n;\W^\ast)\big)\\ &\text{independently for } n=1,\ldots,N.
\end{aligned}
\label{eq:generative}
\end{equation}
}

\begin{table}[t]
\centering\footnotesize
\caption{Prior selection for Bayesian LoRA.}
\label{tab:needs}
\setlength{\tabcolsep}{4pt}
\begin{tabular}{@{}lccc@{}}
\toprule
 & \shortstack{validation\\search} & \shortstack{marginal\\likelihood} & ours\\
\midrule
available before fine-tuning & $\times$ & $\times$ & \checkmark\\
needs no held-out data & $\times$ & \checkmark & \checkmark\\
one estimate for all ranks & $\times$ & $\times$ & \checkmark\\
\bottomrule
\end{tabular}
\end{table}

\paragraph{Prior predictive matching.}
The generative model in \eqref{eq:generative} induces 
the prior predictive distribution
$p(\out \mid \sigma, K) = \int p(\out \mid \W)\, p(\W \mid \sigma, K)\, \mathrm{d}\W$, which describes how
the outputs should scatter around the pretrained function under the chosen prior. 
Then,
prior-predictive
matching \citep{Bayesian-Workflow:2026} can be used to select the hyperparameters so that simple moments of this distribution agree with the
corresponding empirical quantities. 
This, however, requires expressing the predictive distribution as a function of $(\sigma, K)$ (Section~\ref{sec:moments}) and identifying which empirical quantities should be matched (Section~\ref{sec:estimation}).

\section{Prior predictive moments} 
\label{sec:moments}

In this section, we 
find the characteristics of the prior predictive distribution for neural networks with Bayesian LoRA. First, we discuss the moments of the updates $\Delta W$
(Section~\ref{sec:momlemma}) and then, we show how they can be propagated through the network (Section~\ref{sec:propagation}).

\subsection{Moments of the update}
\label{sec:momlemma}

Bayesian LoRA  is an instance of Bayesian matrix factorization (BMF) in latent space. Therefore, the moments derived originally for BMF \citep{dasilva2023prior} can be extended to Bayesian LoRA:


\begin{lemma}[proof in Appendix~\ref{app:proofs}]
\label{lem:moments}
With zero-mean factors of scale $\sigma$, the entries of $\W$ have mean zero, are mutually
uncorrelated, and share the variance $\theta = K\sigma^4$:
\begin{equation}
\Cov\big(\W_{ij}, \W_{i'j'}\big)  = 
\begin{cases} \theta & \text{if } i = i' \text{ and } j = j', \\ 0 & \text{otherwise.} \end{cases}
\end{equation}
Their fourth moment is $3K(K{+}2)\sigma^8$, so the kurtosis equals $3 + 6/K$. \\
Under the general
factor prior \eqref{eq:factorprior} the mean becomes $K\mu_U\mu_V$, the variance becomes
$K\big((\sigma_U^2{+}\mu_U^2)(\sigma_V^2{+}\mu_V^2) - \mu_U^2\mu_V^2\big)$, and distinct
entries covary only within a row (covariance $K\sigma_U^2\mu_V^2$, through the shared row of $U$)
or within a column (covariance $K\mu_U^2\sigma_V^2$, through the shared column of $V$).
\end{lemma}

Uncorrelated entries with a common variance mean that, up to
second order, the prior acts on the network like isotropic noise of strength $\theta$ injected
into the adapted weights. 
The variance $\theta = K\sigma^4$ couples the rank and
the scale into one number, and therefore,
second-order quantities cannot distinguish a rank-4 prior from a
rank-16 prior with a suitably smaller scale. 
To distinguish them one needs to employ   
the
$6/K$ kurtosis term. 
In theory, 
if the rank itself had to be
inferred from data rather than chosen by budget, fourth-order prior-predictive information could be used. 
We verify it by simulation
(Section~\ref{sec:exp-K}). 

\subsection{Propagation through the network}
\label{sec:propagation}

Let $z = \mathrm{vec}(\W) \in \mathbb{R}^P$ denote the update with its $P = N_o N_i$ entries stacked
into one vector.
$g(x) \in \mathbb{R}^P$ is the gradient of the network output with respect to
$z$.
$H(x) \in \mathbb{R}^{P\times P}$ is the
corresponding Hessian, with entries $H_{pq}(x) = \partial^2 \nn / \partial z_p\,\partial z_q$. 
Finally, let $s(x) = \lVert g(x)\rVert^2$ and $t(x) = \Tr H(x)$. 

We show the propagation of moments of the latent variables to observed quantities (i.e., outputs of the network) for a network $\nn(x;\W)$ with a fixed example input $x$. For brevity, we consider just a single scalar output (for example, the probability of a single class). The only difference for multiple outputs is the additional dimension for all the objects:
for example, the Hessian matrix becomes a tensor of Hessian matrices (one for each output).

The second-order Taylor expansion around the 
pretrained network $\nn(x; 0)$ is 
\begin{equation}
\begin{split}
\nn(x;\W) &\;\approx\; \nn(x;0) \;+\; \langle g(x), z\rangle \\
& +\; \tfrac12\, z^\top H(x)\, z.
\label{eq:taylor}
\end{split}
\end{equation}
Under the prior the update is random, so for each fixed input $x$ the left-hand side of
\eqref{eq:taylor} is exactly the random output $\out(x) = \nn(x;\W)$ of
Section~\ref{sec:setting}, and its moments follow by taking expectations of the right-hand side over
the prior, term by term. The linear term vanishes because
$\E[\W] = 0$. The quadratic term is an expectation of a quadratic form in the zero-mean vector
$z$. Written out over pairs of entries it is
$z^\top H z = \sum_{p,q} H_{pq}\, z_p z_q$, and since $H(x)$ is fixed and
$\E[z_p z_q] = \Cov(z_p, z_q)$ for zero-mean entries, $\E[z^\top H z] = \sum_{p,q}
H_{pq}\Cov(z_p, z_q)$. Lemma~\ref{lem:moments} states that, with zero-mean factors, the entries of
$\W$ are mutually uncorrelated and share the variance $\theta$. Since each $z_p$ is one entry of
$\W$, this means $\Cov(z_p, z_q) = \theta$ when $p = q$ and zero otherwise, so only the diagonal
survives and the double sum collapses onto the trace,
$\theta \sum_p H_{pp} = \theta\, t(x)$. Consequently, the mean shift of each output, expected under the prior, is
\begin{equation}
\E[\out(x)] - \nn(x;0) \;\approx\; \tfrac12\,\theta\, t(x).
\label{eq:mean}
\end{equation}

On the other hand, second moments can be obtained by multiplying two copies of \eqref{eq:taylor}, taken at inputs $x$ and
$x'$, and then averaging over the prior again. At leading order only the product of the two linear
terms survives, and it collapses exactly as before:
$\E\big[\langle g(x), z\rangle\,\langle g(x'), z\rangle\big]
= \sum_{pq} g_p(x)\, g_q(x')\, \Cov\big(z_p, z_q\big)
= \theta\,\langle g(x), g(x')\rangle$. The cross terms between a linear and a quadratic factor
contain third moments of $\W$, which vanish by the sign symmetry of the factors, and products of
two quadratic terms are of order $\theta^2$ and dropped (Appendix~\ref{app:proofs}). This yields
the prior predictive variance and the predictive covariance between the outputs at two
different inputs. It is nonzero because both are functions of the same random draw $\W^*$:
\begin{equation}
\Var[\out(x)] \approx \theta\, s(x),
\
\Cov[\out(x), \out(x')] \approx \theta\,\big\langle g(x), g(x')\big\rangle .
\label{eq:varcov}
\end{equation}
Equations \ref{eq:mean} and \ref{eq:varcov} specify the prior predictive distribution. 
The pretrained output at each input $x$ is shifted in proportion to the
local curvature (\eqref{eq:mean}), and the outputs fluctuate jointly around the pretrained function with a covariance
given by \eqref{eq:varcov}. 
Note that the prior predictive distribution depends on its hyperparameters $(\sigma, K)$ entirely through the single scalar~$\theta$. 

Appendix~\ref{app:proofs}
collects the omitted algebra and the study of the higher-order terms.  Section~\ref{sec:exp-valid} studies
regions where the approximation is accurate.

\subsection{Practical consequences}
\label{sec:ranklaw}

The prior predictive distribution 
induced by the prior on $\W$ depends only on $\theta$.
Consequently, the problem of finding the prior's hyperparameters 
reduces to estimation of $\hat \theta$.
However, practical LoRA
implementations conventionally use $r$ rather than $K$, and have the additional fixed scaling hyperparameter $\alpha$, i.e., $\W = \tfrac{\alpha}{r} UV.$ 
The additional multiplier $\alpha/r$ scales the variance from  Lemma~\ref{lem:moments} by $(\alpha/r)^2$, so
\begin{equation}
\theta = \Big(\tfrac{\alpha}{r}\Big)^{2} r\,\sigma^4 = \frac{\alpha^2\sigma^4}{r}
\ \Longrightarrow \
\lhat = \sigma^{-2} = \frac{\alpha}{\sqrt{\hat \theta\, r}} .
\label{eq:lambda}
\end{equation}

\begin{corollary}
\label{cor:rank}
If the data determine an optimal value $\theta^\ast$ of $\theta$, then the optimal prior precision
at rank $r$ obeys $\lambda^\ast(r) \propto r^{-1/2}$ at fixed $\alpha$.
\end{corollary}

Corollary~\ref{cor:rank} is the
Bayesian counterpart of rank-corrected scaling rules for deterministic LoRA
\citep{kalajdzievski2023rslora}.
It implies that any procedure tuning $\lambda$ 
must reproduce this scaling, and a prior-predictive estimate transfers across ranks without a new
search, because $\that$ does not depend on the rank. Section~\ref{sec:exp-rank} tests the
scaling in fine-tuning.

\section{Moment matching}
\label{sec:estimation}

In the standard matrix factorization task every observed matrix entry is a new sample from the prior, so ordinary averaging over the observations estimates prior-predictive moments directly. 
On the other hand, in Bayesian LoRA there is just  \emph{one} draw
$\W^\ast$, evaluated at many inputs and read out through noisy labels, as in \eqref{eq:generative}. 
Therefore, averaging over examples finds the expectations over inputs and over label noise, but not over
the prior itself, making the estimation problem more challenging.

\label{sec:naive}

\subsection{Basics}

We start with a binary classification problem, where the network scalar output
(as specified in Section~\ref{sec:propagation}) is the probability of the positive class. Let $p_{0,n} = \nn(x_n; 0)$, $p_n = \nn(x_n; \W^\ast)$ and
$r_n = y_n - p_{0,n}$. The first expression denotes the pretrained probability at training input $x_n$, a known number obtained by
running the pretrained network. 
The second is the probability under the realized draw. It is not observable because $\W^\ast$
is not, and 
the observed label is
$y_n \mid \W^\ast \sim \mathrm{Bernoulli}(p_n)$, i.e.,
$p_n = \E[y_n \mid \W^\ast]$. 
The third is the observable residual for example $n$. 
\\
From \eqref{eq:mean} we get the expected residual:
\begin{equation}
\begin{split}
\E[r_n] &= \E\big[\E[y_n \mid \W^\ast]\big] - p_{0,n} = \E[p_n] - p_{0,n} \\
&= \E[\out(x_n)] - \nn(x_n; 0)
\approx \tfrac12\,\theta\, t_n ,
\label{eq:perpoint}
\end{split}
\end{equation}
where $\E[p_n] = \E[\out(x_n)]$ because $p_n$ is exactly the random output $\out(x_n)$ (Section~\ref{sec:setting}), evaluated at the prior draw that generated the data.
\\
We also abbreviate 
$g_n = g(x_n)$, $s_n = s(x_n)$,
$t_n = t(x_n)$.

\subsection{Binary classification estimator}
\label{sec:crossexample_covariances}

The most straightforward use of \eqref{eq:perpoint}, suggested by matrix-factorization matching, is
to average both sides over the dataset and solve for $\theta$. This naive aggregate estimator fails
for several reasons, primarily because binary labels satisfy $y^2 = y$, so every statistic computed
from a single example informs only the first moment, which by \eqref{eq:mean} carries just a weak
second-order trace signal. In addition, on roughly balanced tasks the traces $t_n$ change sign and
cancel in the sum. The details are discussed in Appendix~\ref{app:naive}.

On the other hand, by \eqref{eq:varcov},
the prior scale $\sigma$ determines the covariance between the outputs at different
inputs. Since all labels share the single draw $\W^\ast$, the residuals of different examples are correlated, which allows us to estimate $\theta$.
Each residual is the functional shift of the draw plus the label noise,
$r_n = (p_n - p_{0,n}) + \varepsilon_n$ with $\varepsilon_n = y_n - p_n$.
Given $\W^\ast$, the labels are independent with $y_n \mid \W^\ast \sim \mathrm{Bernoulli}(p_n)$,
so the noise terms $\varepsilon_n$ are also independent across examples and satisfy
$\E[\varepsilon_n \mid \W^\ast] = 0$. 
We identify the dependence of the residuals on $\theta$ by taking the expectation of
$r_n r_m$ for two examples $n \neq m$ jointly over the draw and the label noise (as in the model
\eqref{eq:generative}). Conditioning on $\W^\ast$, every term of the product that contains
$\varepsilon_n$ or $\varepsilon_m$ has zero expectation, and by \eqref{eq:varcov}
\begin{equation}
\begin{split}
\E[r_n r_m] & =  \Cov(p_n, p_m) + O(\theta^2) \\
&\;\approx\; \theta\,\langle g_n, g_m\rangle, \qquad n \neq m .
\end{split}
\label{eq:pair}
\end{equation}
Equation~\ref{eq:pair} 
gives one equation for every pair of
examples, and it does not suffer from the problems of the naive estimator (Appendix~\ref{app:naive}). 
We estimate $\theta$ from these equations by least squares. Minimizing
$\sum_{n\neq m}\big(r_n r_m - \theta\,\langle g_n, g_m\rangle\big)^2$ gives
$\that = \sum_{n\neq m} r_n r_m\,\langle g_n, g_m\rangle \big/ \sum_{n\neq m} \langle g_n, g_m\rangle^2$. 

An efficient implementation of the estimator does not
require forming the $N^2$ pairs. Summing over all pairs and subtracting the terms with $n = m$ yields
$\sum_{n\neq m} r_n r_m \langle g_n, g_m\rangle = \lVert \sum_n r_n g_n \rVert^2 - \sum_n r_n^2 s_n$ and
$\sum_{n\neq m} \langle g_n, g_m\rangle^2 = \lVert G^\top G\rVert_F^2 - \sum_n s_n^2$, where
$G \in \mathbb{R}^{N\times P}$ has the per-example gradients $g_n$ as its rows, because
$(G G^\top)_{nm} = \langle g_n, g_m\rangle$ and $\lVert G G^\top\rVert_F = \lVert G^\top G\rVert_F$. Hence, the \emph{cross-covariance estimator} for binary classification tasks is
\begin{equation}
\that  = 
\frac{\big\lVert \sum_n r_n g_n \big\rVert^2 \;-\; \sum_n r_n^2\, s_n}
     {\big\lVert G^\top G\big\rVert_F^2 \;-\; \sum_n s_n^2}\, .
\label{eq:covest}
\end{equation}
Its complexity is linear in the number of examples. The subtracted terms are exactly
the same-example terms, whose observation noise does not cancel.

\subsection{Regression estimator}

For \emph{regression} tasks the likelihood in \eqref{eq:generative} is Gaussian:
$y_n \mid \W^\ast \sim \mathcal N\big(\nn(x_n; \W^\ast), \tau^2\big)$, with observation-noise
variance $\tau^2$. The derivation of \eqref{eq:pair} assumes only that the label noise
$\varepsilon_n$ has zero mean given the draw and is independent across examples. Both also hold for regression. Therefore, \eqref{eq:pair} and \eqref{eq:covest} can be applied directly to the residuals $r_n = y_n - \nn(x_n; 0)$,
and $\tau$ need not be known. If $\tau$ is known, the per-example second moments yield the following
alternative estimator:
\begin{equation}
\E[r_n^2]  \approx  \theta\, s_n + \tau^2
\ \Longrightarrow \
\that_{\mathrm{reg}}  =  \frac{\tfrac1N \sum_n r_n^2 \;-\; \tau^2}{\tfrac1N \sum_n s_n}\,.
\label{eq:regest}
\end{equation}

\subsection{Multiclass estimator}

For \emph{multiclass} models and tasks (with $C$ outputs and $C$ classes), the residuals are vectors with entries
$r_{nc} = \mathbf{1}[y_n = c] - p_{0,c}(x_n)$, the gradients $g_n^{(c)}$ need to be computed per class, and
the pair equations $\E[r_{nc}\, r_{mc'}] \approx \theta\, \langle g_n^{(c)}, g_m^{(c')}\rangle$
range over pairs of example-and-class rows. 
Since the label noise is dependent across the classes of one example, we use only products of
  residuals from different examples.
Let
$M_n \in \mathbb{R}^{C\times P}$ have the $C$ class gradients of example $n$ as its rows, and let
$G$ have all $NC$ of them. Repeating the procedure of the binary estimator, we obtain
\begin{equation}
\that_{\mathrm{mc}}  = 
\frac{\big\lVert \sum_{n,c} r_{nc}\, g_n^{(c)} \big\rVert^2
      \;-\; \sum_n \big\lVert \sum_c r_{nc}\, g_n^{(c)} \big\rVert^2}
     {\big\lVert G^\top G \big\rVert_F^2 \;-\; \sum_n \big\lVert M_n M_n^\top \big\rVert_F^2}\,,
\label{eq:mcest}
\end{equation}
which reduces to \eqref{eq:covest} for $C = 1$. 

\section{Implementation}
\label{sec:implementation}

The estimators in Section~\ref{sec:estimation} require per-example gradients with respect to each adapted weight matrix. Explicitly storing these gradients can be prohibitively expensive for large models, such as Transformer-based MLMs or LLMs, where many high-dimensional matrices are adapted over thousands of examples. We therefore develop an implementation that estimates the required quantities in one pass over the training data, without storing the full gradient matrices.

\paragraph{Multiple adapters.}
LoRA is typically applied to multiple weight matrices throughout the network. Let $\ell=1,\ldots,L$ index the adapted modules. Under independent priors across adapters, their first-order contributions are uncorrelated, and the cross-example covariance decomposes as
\begin{equation}
\E[r_n r_m]
\approx
\sum_{\ell=1}^{L}
\theta_\ell
\big\langle g_n^{(\ell)}, g_m^{(\ell)}\big\rangle,
\qquad n \neq m,
\label{eq:multimod}
\end{equation}
where $g_n^{(\ell)}$ denotes the gradient associated with adapter $\ell$ for example $n$. When a common scale is shared across adapters, as in the standard Laplace-LoRA prior, the inner products can be summed and \eqref{eq:covest} applies directly. Allowing a separate scale for each adapter instead yields an $L$-dimensional least-squares problem, which is dominated by noise at typical single-task sample sizes (Appendix~\ref{app:estimators}). We therefore use a shared scale in our experiments.

\subsection{Gradient sketches}
\label{sec:alg}

To implement the estimator efficiently, we replace the per-example gradients by random sketches. For a linear layer, the per-example gradient in matrix form, $G_n \in \mathbb{R}^{N_o\times N_i}$, is a sum of outer products over token positions, $G_n = \sum_t \delta_{n,t} h_{n,t}^{\top}$, where $h_{n,t}$ is the layer input and $\delta_{n,t}$ is the gradient of the scalar output with respect to the layer output.

Our estimators depend on these gradients only through inner products, which can be preserved in expectation by random projections. Let $S_{\mathrm{out}}$ and $S_{\mathrm{in}}$ be Gaussian sketch matrices with $d$ rows, normalized such that $
\E[S^\top S] = I $.
We replace each gradient by the two-sided sketch
\begin{equation}
\tilde g_n = S_{\mathrm{out}}\, G_n\, S_{\mathrm{in}}^\top = \sum_t
\big(S_{\mathrm{out}}\delta_{n,t}\big)
\big(S_{\mathrm{in}}h_{n,t}\big)^\top.
\label{eq:sketch}
\end{equation}
Because the sketch is linear in $G_n$, it can be accumulated directly from the projected token-level factors without forming the full gradient matrix. Moreover,
$\E\big[\langle \tilde g_n,\tilde g_m\rangle\big] = \langle g_n,g_m\rangle$.

The denominator of \eqref{eq:covest} contains squared inner products. Squaring a single sketched estimate introduces a positive bias due to the sketch variance. We therefore use two independent sketches and multiply the corresponding inner-product estimates, which yields an unbiased estimator of the squared inner product (Appendix~\ref{app:sketch}).

\subsection{Algorithm}

\begin{algorithm}[t]
\caption{Closed-form prior precision for LoRA}
\label{alg:main}
\begin{algorithmic}[1]
\State Residuals $r_n = y_n - \nn(x_n;0)$.
\State One forward and backward pass: independent sketches $\tilde g_n^{(1)}, \tilde g_n^{(2)}$
by \eqref{eq:sketch} per module.
\State $\that$ by \eqref{eq:covest} on the sketches, squared inner products across the two.
\State \Return $\lhat = \alpha/\sqrt{\that\,r}$, or maximal precision if $\that \le 0$.
\end{algorithmic}
\end{algorithm}

Algorithm~\ref{alg:main} requires a single forward-backward pass over the training set. If $\that \leq 0$, the estimated cross-example signal is non-positive and we set the prior precision to its maximum value. The resulting estimate depends only on the pretrained model and the task data, and is therefore available before fine-tuning. It can be used directly as the Bayesian prior precision or as the corresponding regularization strength during optimization (Appendix~\ref{app:train-lambda}). By Corollary~\ref{cor:rank}, the same computation applies to all LoRA ranks. 

\section{Related work}
\label{sec:related}

Our work relates to Bayesian treatments of LoRA, prior predictive analyses of neural networks, prior
predictive matching, and the selection of the prior precision for Laplace approximations.

\textbf{Bayesian LoRA.} Bayesian treatments of LoRA differ mainly in how they approximate the
posterior over the adapter weights. In particular, there exist variants based on a post-hoc Laplace approximation \citep{yang2024laplacelora},
variational inference during fine-tuning \citep{wang2024blob}, stochastic weight averaging
\citep{onal2024gaussian}, and ensembles of adapters \citep{wang2023lora,doan2025bayesianlowranklearningbella}.
In all
of them, the prior scale is either fixed in advance or tuned after training.
Hierarchical and meta-learned
priors \citep{eide2024bora,kim2025lift,zhang2025abmll} instead learn a shared prior from many related tasks.\\
Our
estimate is complementary: it sets the prior scale of a single task from its own data, before
fine-tuning, and, in principle, can be combined with any of these posterior approximations. 
However, there also exist some other variants of
Bayesian LoRA using a slightly different structure of the update, which places the posterior on a small matrix
inside a projected low-rank subspace rather than on the projection matrices
\citep{marszalek-etal-2025-minimal,dubovik2026bayesian}. Our
derivations can be adapted to these structures with few changes, but this is beyond the scope of
this paper.

\textbf{Prior predictive analysis of neural networks.} The distribution of the outputs of standard networks under random weights has been studied at initialization and in the infinite-width limit.
In this limit it becomes a Gaussian process \citep{neal1996bayesian,lee2018deep}, and training of the linearized
network is governed by the neural tangent kernel \citep{jacot2018neural}. At finite width, the prior predictive is known in closed form and becomes increasingly heavy-tailed as depth grows relative to
width \citep{noci2021precise,zavatoneveth2021exact}. It is analogous to the excess kurtosis $6/K$ of a
rank-$K$ update (Lemma~\ref{lem:moments}).
\\
Moments of the outputs under random weights also guide practical choices of scale. For example,
signal propagation analyses use them to choose the initialization scale at which deep networks remain trainable \citep{schoenholz2017deep}.
\citet{noci2022signal} extend this analysis to transformers. These analyses choose a scale as the depth grows, whereas rank-stabilized LoRA applies a related
argument to the rank. It scales
the adapter by $\alpha/\sqrt{r}$ so that the size of its output and of its gradients does not change
with the rank during fine-tuning \citep{kalajdzievski2023rslora}. However, to our knowledge, no
such rule existed for the prior precision of Bayesian LoRA. Corollary~\ref{cor:rank} provides the first
one: it states how the precision must change with the rank so that the prior predictive stays fixed.

Function-space methods specify it either directly as a stochastic process \citep{sun2019functional} or by choosing the weight prior so that the
functions it induces match a target process \citep{tran2022all,pmlr-v286-sendera25a}. Meanwhile, \citet{fortuin2021bayesian} empirically studied
which weight priors suit trained networks.\\
In contrast, we propagate a factorized prior on the update through a second-order expansion around the \emph{pretrained} weights, at finite width, and use the resulting moments to estimate the prior
scale from data.

\textbf{Prior predictive matching.} Several approaches specify the prior through the predictions it
implies. Predictive complexity priors define the prior by comparing the predictions of the model with
those of a simpler reference model \citep{nalisnick2021predictive}. Prior predictive matching instead
chooses the hyperparameters so that statistics of the prior predictive distribution agree with target
values. It connects prior elicitation, which translates knowledge about observable quantities into
prior distributions \citep{mikkola2024prior}, with prior predictive checks, a standard step of the
Bayesian workflow \citep{gabry2019visualization}. In elicitation, the targets are judgments of an
expert, and \citet{hartmann2020flexible} fit them by stochastic optimization, which, however, works only for simple shallow models.
By
contrast, for BMF, \citet{dasilva2023prior} obtain the targets from the data and match prior predictive
moments of the observed entries in closed form. \citet{silva2026on} study when tensor ranks are
identifiable from such moments. Closed-form empirical Bayes solutions also exist for fully observed
variational matrix factorization \citep{nakajima2013global}.\\
Bayesian LoRA, however, differs in that the observations depend on the update only through the nonlinear pretrained network and share a single draw of it (Section~\ref{sec:estimation}).

\textbf{Precision selection and calibration.} The standard ways to choose the prior precision of a
Laplace approximation are maximization of the marginal likelihood
\citep{mackay1992practical,immer2021scalable,antoran2022adapting} and validation search. Both
require a trained (in our case: already a fine-tuned) model.\\
In contrast, our estimate needs neither a trained model nor validation data.
However, when validation data are available, it can additionally use them to fit a temperature
\citep{guo2017calibration}. This optional step allows it to match validation search in test NLL 
(Section~\ref{sec:exp-mlm}).

\section{Experiments}
\label{sec:experiments}

In this Section,
we first validate the theoretical approximations and parameter recovery in controlled experiments, where the data-generating prior is known
(Sections~\ref{sec:exp-valid} to~\ref{sec:exp-K}). 
Then, we evaluate the estimated prior precision on RoBERTa-large, considering predictive performance, calibration, the dependence on the LoRA rank, and out-of-distribution detection
(Sections~\ref{sec:exp-mlm} to~\ref{sec:exp-ood}).

In the controlled experiments, $\W = UV$ is drawn from the factor prior (\eqref{eq:factorprior}), and predictions are computed using a full nonlinear network. 
We parameterize the strength of the induced functional perturbation by $\theta\bar s$, where $\bar s$ is the average of $s(x)$ over the inputs. By \eqref{eq:varcov}, $\theta\bar s$ is the average prior variance of the output.

\subsection{Accuracy of the second-order approximation}
\label{sec:exp-valid}

\begin{figure}[t]
\centering
\includegraphics[width=\columnwidth]{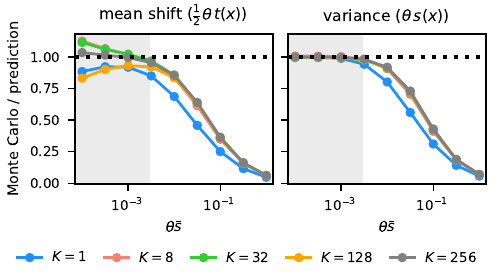}
\caption{Ratio of moments of the network output to their second-order predictions (Equations~\ref{eq:mean} and~\ref{eq:varcov}). Shaded: $\theta\bar s \le 3\cdot10^{-3}$ (the range used in Section~\ref{sec:exp-recover}).}
\label{fig:second}
\end{figure}

We first evaluate the quality of the moment approximations (\eqref{eq:mean} and \eqref{eq:varcov}), on
which our estimators depend. Figure~\ref{fig:second} compares them against Monte Carlo
estimates for a small tanh network (Appendix~\ref{app:toy}).

The approximation is accurate for $\theta\bar s \lesssim 3\cdot10^{-3}$. In this range,
the variance error is below $2\%$ for $K\geq8$, while the mean-shift error is at most
$17\%$. At fixed $\theta$, the results are nearly invariant to $K$, as predicted in
Section~\ref{sec:propagation}. The main exception is $K=1$, where the first neglected
term depends on the kurtosis $3+6/K$ (Appendix~\ref{app:proofs}).

For stronger priors, the second-order expansion overestimates both moments as the nonlinear output saturates. 
The discrepancy becomes substantial beyond $\theta\bar s\approx3\cdot10^{-3}$ and exceeds $50\%$ at $10^{-1}$. Since the estimator inverts these moments, this regime produces an upward bias in $\lhat$. We therefore restrict the experiment in Section~\ref{sec:exp-recover} (scale recovery) to the regime where the approximation is accurate.

\subsection{Recovery of the prior scale}
\label{sec:exp-recover}
\begin{figure}[t]
\centering
\includegraphics[width=\columnwidth]{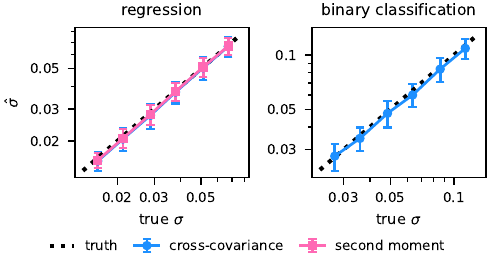}
\caption{Recovery of the prior scale: average $\hat\sigma$ $\pm$ one standard deviation.}
\label{fig:sigma}
\end{figure}

We next test whether the cross-covariance estimator recovers the prior scale from a
single task (Figure~\ref{fig:sigma}). Each task draws one update $\W^\ast$ at the true
$\sigma$, generates $N$ labels according to \eqref{eq:generative}, and estimates
$\hat\sigma=(\that/K)^{1/4}$ with $K=8$. We consider regression with Gaussian 
noise and binary classification with a logistic model.
For regression, we additionally
compare \eqref{eq:covest} with \eqref{eq:regest} (which is given the true noise level).
The prior scales span $\theta\bar s$ from $10^{-5}$ to $3\cdot10^{-3}$, within the
regime validated in Section~\ref{sec:exp-valid}. Appendix~\ref{app:toy} covers any
remaining details.

Across tasks with positive estimates, $\hat\sigma$ is within $4\%$ of the true scale
for regression and within $6\%$ for classification. The cross-covariance estimator,
which does not use $\tau$, is as accurate as the second-moment estimator that uses
the true noise level. At the smallest classification scale, $5\%$ of tasks yield
$\that\leq0$. These cases are excluded. 
The standard
deviation of $\hat\sigma$ across tasks is $11$ to $17\%$ of $\sigma$, compared with
$9$ to $13\%$ for the second-moment estimator. 
This spread comes mostly from the update itself: each task observes a single draw $\W^\ast$, whose entries are by chance larger or smaller than the prior scale implies (Section~\ref{sec:estimation}).

\subsection{Rank information in fourth moments}
\label{sec:exp-K}

Second-order moments depend on the rank and prior scale only through $\theta$ and
therefore cannot identify $K$ separately (Section~\ref{sec:momlemma}). 
In this subsection, however, we verify that
rank information appears in the fourth moment of the prior predictive.

To first order, the output perturbation for example $n$ is
$\langle G_n,\W\rangle$, where $G_n\in\mathbb{R}^{N_o\times N_i}$ is the matrix form
of $g(x_n)$. Its kurtosis is
\begin{equation}
\kappa_n = 3 + \frac{6\rho_n}{K},
\qquad
\rho_n = \frac{\sum_i \gamma_{ni}^4}{\big(\sum_i \gamma_{ni}^2\big)^2},
\label{eq:kurt}
\end{equation}
where $\gamma_{ni}$ are the singular values of $G_n$
(Appendix~\ref{app:kurtosis}). For a single entry of $\W$, $\rho_n=1$, recovering the
$3+6/K$ result of Lemma~\ref{lem:moments}. We estimate the slope $b$ of
$\kappa_n-3$ against $\rho_n$ and set $\hat K=6/b$.

The networks contain one tanh layer followed by a linear output for regression and a
logistic output for classification. Each of 24 examples averages two inputs, so that
$G_n$ has rank two (Appendix~\ref{app:toy}). We estimate the kurtosis from
$10{,}000$ independent draws of $\W$ without label noise. This experiment therefore
tests identifiability under the prior predictive. It does not estimate $K$ from a
single task.

For the linearized output, the median $\hat K$ is within $13\%$ of the true rank.
At $\theta\bar s=2\cdot10^{-4}$, the nonlinear network agrees with this estimate
within $3\%$. Recovery deteriorates for stronger priors, particularly for
classification. This follows from \eqref{eq:kurt}: the rank enters only through the
small correction $6\rho_n/K$, making fourth moments more sensitive to the nonlinearity than the
second-order quantities used for scale estimation. The experiment thus confirms that
the prior predictive contains rank information, but also that this information is
fragile under nonlinear distortion.

\begin{figure}[t]
\centering
\includegraphics[width=\columnwidth]{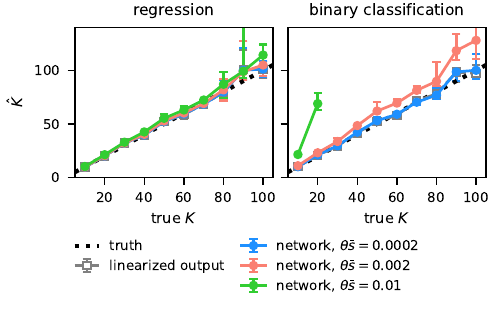}
\caption{Recovery of the rank from the kurtosis (Equation~\ref{eq:kurt}). Median $\hat K$ over 10
repetitions, with interquartile bars. The green line in the right panel ends early, as for larger $K$
the estimates $\hat K$ are  undefined or outside the plotted range.
}
\label{fig:rank}
\end{figure}

\subsection{Predictive performance}
\label{sec:exp-mlm}

We now turn to Bayesian LoRA fine-tuning of RoBERTa-large \citep{liu2019roberta}, where the true prior is unknown and the estimated precision is evaluated by its predictive performance. We follow Laplace-LoRA \citep{yang2024laplacelora} and use the pretrained masked-language-model head for classification, so that $\nn(x;0)$ defines
task predictions before adaptation. We consider SST-2, MR, CR, and AG News, with 64 to $4{,}096$ training examples and three seeds.

The posterior is a Kronecker-factored Laplace approximation with prior precision $\lambda$ \citep{daxberger2021laplace}, and predictions use the linearized predictive \citep{immer2021improving}. As is standard for deep networks, the posterior is fitted (in stage 2) after point-wise fine-tuning (=stage 1), so $\lambda$ enters only after training and all methods share the point-wise fine-tuned model and its Laplace approximation, differing only in $\lambda$. However, note that our method finds $\lambda=\lhat$ already before fine-tuning, which also allows its use during 1st stage of training (we study it in Appendix~\ref{app:train-lambda}). We compare against the MAP predictor (the fine-tuned weights alone, i.e., $\lambda = 10^8$, which makes the posterior variance negligible) and two post-training choices used by Laplace-LoRA: maximization of the marginal likelihood and selection of $\lambda$ by validation NLL. The latter reserves $20\%$ of the available examples, and henceforth all methods are trained only on the remaining $80\%$. Appendix~\ref{app:mlm} gives the full protocol and per-task results (Tables~\ref{tab:mlm-agnews} to~\ref{tab:mlm-sst2}).

\begin{figure}[t]
\centering
\includegraphics[width=\columnwidth]{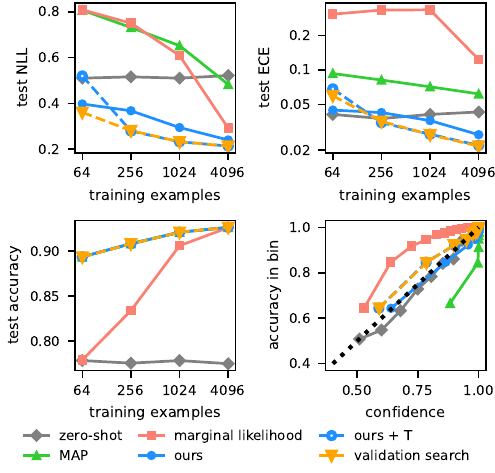}
\caption{RoBERTa-large on SST-2, MR, CR and AG News, mean over the four tasks and three seeds.
Bottom right: reliability diagram at $4{,}096$ training examples, ten bins of equal size. Solid
lines use no validation data. Ours + T adds a temperature fitted on the validation set. The other
methods with a temperature, and each task separately, are in Figure~\ref{fig:mlm-tasks}.}
\label{fig:mlm}
\end{figure}

Among methods that do not use the validation set for precision selection, ours achieves
the lowest test NLL at every training-set size (Figure~\ref{fig:mlm}). It improves over
MAP in all 46 runs and over marginal likelihood in all 35 runs up to $1{,}024$
examples. At 64 examples, it is also the only one of these methods to improve over the
pretrained model, signifying its robustness for tiny data set sizes.
Its ECE is lower than that of MAP and marginal likelihood while its
accuracy matches MAP. 
Finally, we note that the prior estimate takes at most 40 seconds, compared with
30-40 minutes for LoRA fine-tuning.

Validation search achieves a further NLL reduction of $0.03$-$0.09$ nats. 
However, when
validation data are available, fitting a temperature to our predictive distribution
(ours + T) reduces this gap to at most $0.01$ nats from 256 examples onward. At
64 examples, where the validation set contains only 13 observations, temperature
scaling instead increases NLL and ECE for all methods considered.

\subsection{Prior precision across LoRA ranks}
\label{sec:exp-rank}

By Corollary~\ref{cor:rank}, one
estimate $\that$ determines the prior at every rank, with $\lhat \propto r^{-1/2}$, so the rank can be
changed without estimating again. 

We test this by fine-tuning on AG News and CR with $1{,}024$ training examples,
LoRA ranks from 4 to 64, and three seeds, with the posterior and predictive of
Section~\ref{sec:exp-mlm}. As a reference, $\lambda^\ast$ is the precision with the lowest validation
NLL among values from $10^{-2}$ to $10^6$ (Appendix~\ref{app:rank}).

\begin{figure}[t]
\centering
\begin{subfigure}{\columnwidth}
\centering
\includegraphics[width=\columnwidth]{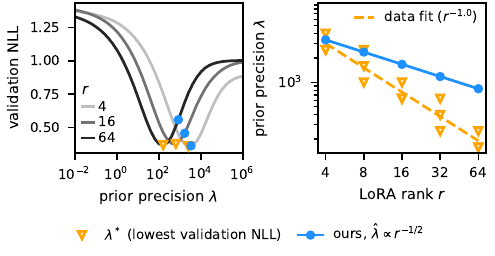}
\caption{AG News}
\end{subfigure}
\par\medskip
\begin{subfigure}{\columnwidth}
\centering
\includegraphics[width=\columnwidth]{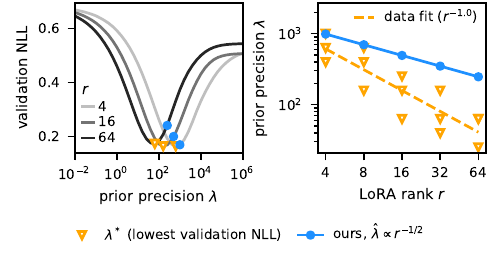}
\caption{CR}
\end{subfigure}
\caption{Prior precision across LoRA ranks on RoBERTa-large. Left: validation NLL against the prior
precision at three ranks for one seed, with its minimum $\lambda^\ast$ and ours, $\lhat$. Right:
$\lambda^\ast$ of every run against the rank, its least-squares fit, and ours (medians).}
\label{fig:rank-transfer}
\end{figure}

Figure~\ref{fig:rank-transfer} shows that
our $\lhat$ is close to $\lambda^\ast$ at small ranks, but it diverges from the optimum at higher ranks, and its NLL grows with the rank. 
This is because
$\lambda^\ast$ falls faster than $r^{-1/2}$, close to $r^{-1}$ in both tasks (fitted slope $-0.99$).
The faster decrease does not
contradict the corollary, which holds for the prior. 
It arises from a practical choice, the linearized
predictive used for prediction, whose variance scales as $1/(r\lambda)$ rather than as
$1/(r\lambda^2)$ under the prior (Appendix~\ref{app:linearized}).
Finally, we also note that
a
temperature fitted on the validation set removes this loss, and ours + T matches validation search
with a temperature at every rank (Table~\ref{tab:rank-large}). 

\subsection{Detection of out-of-distribution inputs}
\label{sec:exp-ood}

\begin{figure}[t]
\centering
\includegraphics[width=\columnwidth]{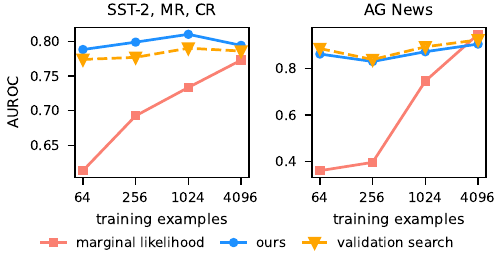}
\caption{AUROC for separating OOD inputs from the task's own test inputs, with the mutual
information as the score, mean over three seeds (left, also over the three tasks). The OOD inputs are AG News for the
sentiment tasks and SST-2 for AG News. Per-task results are in Figure~\ref{fig:ood-tasks} and
Table~\ref{tab:mlm-ood}.}
\label{fig:ood}
\end{figure}

Finally, we evaluate posterior uncertainty on out-of-distribution (OOD) inputs
(Figure~\ref{fig:ood}). For each test input, the linearized predictive gives a Gaussian
distribution over the logits (Appendix~\ref{app:kfac}). We sample the logits from it, which needs no
further passes through the network, and use the mutual information between the prediction and the
weights as the uncertainty score. We report the area under the ROC curve (AUROC) for distinguishing the task's test inputs from
those of another task: AG News is OOD for the sentiment tasks, and SST-2 is OOD for
AG News. The MAP predictor has no posterior and is therefore excluded.

For the three sentiment tasks, our method obtains the highest AUROC at every
training-set size, ranging from $0.79$ to $0.81$. It exceeds validation search in
30 of 34 runs and marginal likelihood in all 34, with the largest gain over marginal
likelihood at 64 examples. On AG News, our method obtains AUROC values from $0.83$ to
$0.90$, while validation search is higher by $0.01$-$0.02$. Marginal likelihood is
substantially worse in the low-data regime but reaches the highest AUROC at $4{,}096$
examples. Overall, the estimated prior yields competitive OOD uncertainty without
using validation data for precision selection.

\section{Conclusion}
\label{sec:discussion}

We studied the prior predictive distribution for Bayesian LoRA and introduced a closed-form, moment-based method that specifies the prior. The method does not require a validation set and can be applied before fine-tuning. 
\\
Using a second-order expansion, we showed that the rank $K$ and the factor scale $\sigma$ affect the prior-predictive variance only through $\theta=K\sigma^4$. 
Since all examples of a dataset share the same LoRA update, $\theta$ is reflected in the covariance between residuals of different examples. Matching these covariances yields an estimator of $\theta$, and therefore of the prior precision. Under standard LoRA scaling, this also gives the rank dependence $\lambda\propto r^{-1/2}$.

Controlled experiments show that the second-order approximation is accurate for sufficiently small prior perturbations and that the proposed estimator recovers the prior scale from a single task. 
We further showed that fourth-order moments allow us to recover information about the rank. 
On RoBERTa-large, without validation data, the estimated prior improves NLL and calibration over MAP and marginal-likelihood selection and provides competitive OOD detection. 
Validation search remains slightly better in NLL, but when validation data are available, temperature scaling largely closes this gap.

\paragraph{Limitations and future work.}
The proposed estimator relies on a local second-order approximation and becomes biased when the LoRA update induces large changes in the network output. Extending the method beyond this local regime, for example through higher-order expansions or alternative approximations to the prior predictive, is a possible direction for future work. 
\\
The accuracy of the estimator also varies across tasks because each task provides only one realization of the LoRA update. More efficient estimators that exploit additional structure across examples, tasks, or repeated adaptations could reduce this variability.
\\
The second-order estimator identifies the combined scale $\theta$ but not the rank itself, which must therefore still be chosen separately. Our fourth-order analysis shows that the prior predictive contains information about the rank, but turning this result into a stable estimator from a single task remains open. 

Finally, the approach may extend to generative objectives, where next-token prediction is a multiclass problem at every position, as well as to regression tasks at scale and to settings without an immediately available task-aligned output. Such extensions are natural next steps but are beyond the scope of this work.

\newpage


\bibliography{refs}


\clearpage
\appendix
\thispagestyle{empty}

\onecolumn
\aistatstitle{Prior Predictive Matching for Bayesian LoRA
\\
Supplementary Materials
}

Appendix~\ref{app:proofs} proves the results of
Section~\ref{sec:moments}, and Appendix~\ref{app:estimators} derives further estimators, among them
the naive aggregate and the trace-instrument estimators. Appendix~\ref{app:sketch} presents the
guarantees of the gradient sketches, and Appendix~\ref{app:kfac} the posterior approximations, the
predictive and the evidence. Appendix~\ref{app:toy} describes the settings of the toy
experiments, Appendix~\ref{app:rank} the protocol and full results of the rank study of
Section~\ref{sec:exp-rank}, Appendix~\ref{app:linearized} explains why the best precision for the linearized
predictive scales differently with the rank than under the prior, and Appendix~\ref{app:mlm} the setup and
per-task results of the RoBERTa-large experiments.
Appendix~\ref{app:train-lambda} fine-tunes RoBERTa-large under the estimated prior.

\section{Proofs}
\label{app:proofs}

\subsection{Proof of Lemma~\ref{lem:moments}}
\label{app:prooflemma}

Fix an entry $(i,j)$ and write the summands as $P_k = U_{ik} V_{kj}$, so that
$\W_{ij} = \sum_{k=1}^{K} P_k$. Under \eqref{eq:factorprior} all factor entries are mutually
independent, so the $P_k$ are independent across $k$, and within each $P_k$ the two factors are
independent of one another. We first consider the zero-mean case
$\mu_U = \mu_V = 0$, $\sigma_U = \sigma_V = \sigma$, then the general one.

\textbf{Mean.} Independence of the two factors inside each product yields
\begin{equation}
\E[\W_{ij}]  =  \sum_{k=1}^{K} \E[U_{ik}]\,\E[V_{kj}]  =  \sum_{k=1}^{K} 0\cdot 0  =  0 .
\end{equation}

\textbf{Variance.} Each product has second moment
\begin{equation}
\E[P_k^2]  =  \E[U_{ik}^2]\,\E[V_{kj}^2]  =  \sigma^2 \cdot \sigma^2  =  \sigma^4 ,
\end{equation}
and since the $P_k$ are independent with mean zero, their variances add:
\begin{equation}
\Var(\W_{ij})  =  \sum_{k=1}^{K} \Var(P_k)  =  K\sigma^4  =  \theta .
\end{equation}

\textbf{Distinct entries are uncorrelated.} For $(i,j) \neq (i',j')$, zero means allow us to compute
the covariance as a raw second moment, and independence of the $U$ block from the $V$ block
factorizes each term:
\begin{equation}
\Cov\big(\W_{ij}, \W_{i'j'}\big)  =  \E\big[\W_{ij}\,\W_{i'j'}\big]
 =  \sum_{k=1}^{K}\sum_{k'=1}^{K} \E\big[U_{ik}\,U_{i'k'}\big]\;\E\big[V_{kj}\,V_{k'j'}\big] .
\label{eq:covsum}
\end{equation}
We distinguish three cases, which together cover all pairs. If the entries share a row, $i = i'$ and $j \neq j'$, then
$V_{kj}$ and $V_{k'j'}$ are \emph{distinct} entries of $V$ for every $(k,k')$, hence independent,
so $\E[V_{kj}V_{k'j'}] = \E[V_{kj}]\,\E[V_{k'j'}] = 0$ and every term of \eqref{eq:covsum}
vanishes. If the entries share a column, the same argument applies to the $U$ expectations.
If they share neither index, both expectations factor into products of zero means. In all three cases
the covariance is zero.

\textbf{Fourth moment.} Gaussian factors satisfy $\E[U_{ik}^4] = 3\sigma^4$, so each product has
\begin{equation}
\E[P_k^4]  =  \E[U_{ik}^4]\,\E[V_{kj}^4]  =  3\sigma^4 \cdot 3\sigma^4  =  9\sigma^8 .
\end{equation}
Expand $\big(\sum_k P_k\big)^4$ by the multinomial theorem and compute the expectations term by term.
Each $P_k$ is symmetric about zero and independent of the others, so any term containing some
$P_k$ to an odd power has expectation zero. Only two types of terms have nonzero expectation: the $K$ pure fourth
powers, and the mixed squares $P_k^2 P_l^2$ with $k \neq l$, each carrying the multinomial
coefficient $\binom{4}{2} = 6$ over the $\tfrac12 K(K-1)$ unordered pairs. Hence
\begin{equation}
\E[\W_{ij}^4]  =  K\,\E[P^4] \;+\; 6\cdot\tfrac12 K(K-1)\,\E[P^2]^2
 =  9K\sigma^8 + 3K(K-1)\sigma^8  =  3K(K+2)\,\sigma^8 ,
\end{equation}
and dividing by the squared variance yields the kurtosis
\begin{equation}
\frac{\E[\W_{ij}^4]}{\Var(\W_{ij})^2}  =  \frac{3K(K+2)\sigma^8}{K^2\sigma^8}
 =  \frac{3(K+2)}{K}  =  3 + \frac{6}{K} .
\end{equation}

\textbf{General factor means.} Now consider \eqref{eq:factorprior} with arbitrary
$(\mu_U, \sigma_U, \mu_V, \sigma_V)$. The mean follows directly:
$\E[\W_{ij}] = \sum_k \mu_U\mu_V = K\mu_U\mu_V$. All remaining computations reduce to the pair
moments of factor entries, which follow from independence of distinct entries:
\begin{equation}
\E[U_{ik}U_{ik'}] = \sigma_U^2\,\delta_{kk'} + \mu_U^2 , \qquad
\E[V_{kj}V_{k'j}] = \sigma_V^2\,\delta_{kk'} + \mu_V^2 , \qquad
\E[V_{kj}V_{k'j'}] = \mu_V^2 \;\;\text{for } j \neq j' .
\label{eq:pairmoments}
\end{equation}
For the variance, insert \eqref{eq:pairmoments} into the double sum for $\E[\W_{ij}^2]$ and split
it into diagonal ($k = k'$) and off-diagonal parts:
\begin{equation}
\E[\W_{ij}^2] = \sum_{k,k'} \big(\sigma_U^2\delta_{kk'} + \mu_U^2\big)
\big(\sigma_V^2\delta_{kk'} + \mu_V^2\big)
= K\big(\sigma_U^2 + \mu_U^2\big)\big(\sigma_V^2 + \mu_V^2\big) + K(K-1)\,\mu_U^2\mu_V^2 ,
\end{equation}
and subtract $\E[\W_{ij}]^2 = K^2\mu_U^2\mu_V^2$ to obtain
$\Var(\W_{ij}) = K\big[(\sigma_U^2+\mu_U^2)(\sigma_V^2+\mu_V^2) - \mu_U^2\mu_V^2\big]$ as stated.
For two entries in the same row ($i = i'$, $j \neq j'$), the $V$ expectation is the constant
$\mu_V^2$ by \eqref{eq:pairmoments}, so
\begin{equation}
\E[\W_{ij}\W_{ij'}] = \sum_{k,k'} \big(\sigma_U^2\delta_{kk'} + \mu_U^2\big)\,\mu_V^2
= K\sigma_U^2\mu_V^2 + K^2\mu_U^2\mu_V^2 ,
\end{equation}
and subtracting the product of the means, $K^2\mu_U^2\mu_V^2$, leaves
$\Cov(\W_{ij}, \W_{ij'}) = K\sigma_U^2\mu_V^2$: the shared row of $U$ is the only surviving
source of correlation. The same-column case is the mirror image and yields $K\mu_U^2\sigma_V^2$.

\subsection{From \eqref{eq:taylor} to \eqref{eq:mean} and \eqref{eq:varcov}}
\label{app:proofprop}

Abbreviate $z = \mathrm{vec}(\W)$, $f_0 = \nn(x;0)$, $g = g(x)$, and $H = H(x)$, with primes for
evaluation at $x'$ and indices $p, q, r, s$ running over the entries of $z$ as in
Section~\ref{sec:propagation}, so that the expansion \eqref{eq:taylor} reads
\begin{equation}
\out(x) - f_0 \;\approx\; \langle g, z\rangle \;+\; \tfrac12\, z^\top H z .
\label{eq:taylorz}
\end{equation}
The derivation uses two facts about $z$ from Lemma~\ref{lem:moments}: its entries have
$\E[z_p] = 0$ and $\E[z_p z_q] = \theta\,\delta_{pq}$, with $\delta_{pq}$ the Kronecker delta.

\textbf{Mean.} Compute the expectation of \eqref{eq:taylorz} term by term. The linear term is
$\E\langle g, z\rangle = \langle g, \E z\rangle = 0$. The quadratic term expands as a double sum
over entries,
\begin{equation}
\E\big[z^\top H z\big]  =  \sum_{p,q} H_{pq}\,\E[z_p z_q]
 =  \sum_{p,q} H_{pq}\,\theta\,\delta_{pq}  =  \theta \sum_p H_{pp}  =  \theta\,\Tr H ,
\end{equation}
which, multiplied by $\tfrac12$, yields the mean shift \eqref{eq:mean}.

\textbf{Second moments.} Multiply \eqref{eq:taylorz} by its copy at $x'$ and compute the expectations.
The product splits into four groups of terms:
\begin{align}
\E\big[(\out(x)-f_0)(\out(x')-f_0')\big]
 = & \;\E\big[\langle g, z\rangle\,\langle g', z\rangle\big] \nonumber\\
&+ \tfrac12\,\E\big[\langle g, z\rangle\; z^\top H' z\big]
 \;+\; \tfrac12\,\E\big[\langle g', z\rangle\; z^\top H z\big] \nonumber\\
&+ \tfrac14\,\E\big[\big(z^\top H z\big)\big(z^\top H' z\big)\big] .
\label{eq:prodterms}
\end{align}
As for the mean, the first line simplifies through the entry covariances:
\begin{equation}
\E\big[\langle g, z\rangle\,\langle g', z\rangle\big]
 =  \sum_{p,q} g_p\, g'_q\, \E[z_p z_q]
 =  \theta \sum_p g_p\, g'_p
 =  \theta\,\langle g, g'\rangle .
\end{equation}
The second line consists of third moments $\E[z_p z_q z_r]$. Flipping the sign of $U$ maps
$z \to -z$ while leaving the zero-mean prior \eqref{eq:factorprior} unchanged, so the
distribution of $z$ is symmetric, every odd moment vanishes, and the whole line is zero. The
third line is a fourth-moment expression, $\sum_{p,q,r,s} H_{pq} H'_{rs}\,\E[z_p z_q z_r z_s]$.
Each fourth moment of the entries of $\W$ is a sum of $O(K^2)$ nonzero products of factor
moments, each of order $\sigma^8$, so it is $O(\theta^2)$, and so is the whole line. Passing from the raw second moment to the covariance subtracts the
product of the two mean shifts,
$\big(\tfrac12\theta\, t(x)\big)\big(\tfrac12\theta\, t(x')\big) = O(\theta^2)$, which changes
nothing at leading order. Altogether,
\begin{equation}
\Cov\big[\out(x), \out(x')\big]  =  \theta\,\langle g(x), g(x')\rangle \;+\; O(\theta^2),
\end{equation}
which is the covariance half of \eqref{eq:varcov}, and setting $x' = x$ yields the variance half,
$\Var[\out(x)] = \theta\, s(x) + O(\theta^2)$.

\textbf{The dropped terms.} The $O(\theta^2)$ terms come from the fourth moments of $z$. The
entries of $\W$ are uncorrelated but not independent, because entries in the same row share a row
of $U$ and entries in the same column share a column of $V$. Entries that share neither a row nor
a column behave as independent, $\E[\W_{ij}^2\W_{i'j'}^2] = \theta^2$. Entries that coincide or
share a row or a column do not: $\E[\W_{ij}^4] = 3\theta^2(1+2/K)$,
$\E[\W_{ij}^2\W_{ij'}^2] = \theta^2(1+2/K)$, and the four corners of a rectangle give
$\E[\W_{ij}\W_{ij'}\W_{i'j}\W_{i'j'}] = \theta^2/K$. These corrections of relative order $1/K$
are where the rank first enters beyond $\theta$. This is the same fourth-order separation of $(\sigma, K)$ discussed in
Section~\ref{sec:momlemma}, and it is why every second-order prediction in the main text is a
function of $\theta$ alone.

\subsection{Kurtosis of a linearized output}
\label{app:kurtosis}

Let $G_n \in \mathbb{R}^{N_o\times N_i}$ be a fixed matrix with singular values $\gamma_{ni}$, and let
$\xi = \langle G_n, \W\rangle$ with $\W = UV$ drawn from the zero-mean factor prior. With $u_k$ the
$k$-th column of $U$ and $v_k$ the $k$-th row of $V$,
\begin{equation*}
\xi = \sum_{k=1}^{K} u_k^\top G_n\, v_k ,
\end{equation*}
a sum of $K$ independent and identically distributed terms. The entries of $u_k$ and $v_k$ are
independent $\mathcal N(0, \sigma^2)$, so their distribution does not change under rotations, and in
the singular bases of $G_n$ one term has the distribution of $\sigma^2 \sum_i \gamma_{ni} a_i b_i$ with
$a_i, b_i$ independent standard normal. Its second moment is $\sigma^4 \sum_i \gamma_{ni}^2$. In its
fourth moment the terms with an odd power of some $a_i$ or $b_i$ vanish, and
$\E[a_i^4]\,\E[b_i^4] = 9$, so the fourth moment is
\begin{equation*}
\sigma^8\Big(9\sum_i \gamma_{ni}^4 + 3\sum_{i\neq j}\gamma_{ni}^2\gamma_{nj}^2\Big)
= \sigma^8\Big(6\sum_i \gamma_{ni}^4 + 3\big(\textstyle\sum_i \gamma_{ni}^2\big)^2\Big).
\end{equation*}
One term therefore has kurtosis $3 + 6\rho_n$ with $\rho_n = \sum_i\gamma_{ni}^4 / (\sum_i\gamma_{ni}^2)^2$.
For a sum of $K$ independent and identically distributed terms with mean zero, the excess kurtosis
is divided by $K$, which yields \eqref{eq:kurt}. The second moment of $\xi$ is
$K\sigma^4\sum_i\gamma_{ni}^2 = \theta\lVert G_n\rVert_F^2$, in agreement with \eqref{eq:varcov}.

\section{Estimator instances and derivations}
\label{app:estimators}

\paragraph{Pair equation and least squares.} The derivation of the cross-covariance estimator
\eqref{eq:covest} from the pair equation \eqref{eq:pair}, including the least-squares step and
the two add-and-subtract identities, appears in full in Section~\ref{sec:estimation}. This
appendix collects the instances and variants that the main text states without proof.

\paragraph{The naive aggregate estimator and why it fails.}
\label{app:naive}
The most direct use of \eqref{eq:perpoint} is the one that matrix-factorization matching would
suggest: average both sides over the dataset and solve for the single unknown, which yields the
\emph{naive aggregate estimator}
\begin{equation}
\that_{\mathrm{naive}}  =  \frac{2\sum_n r_n}{\sum_n t_n}\, .
\label{eq:naive}
\end{equation}
It matches the average observed label against the average predicted output, as one matches
the average observed matrix entry in factorization models. It fails here for two separate reasons.

\begin{figure}[t]
\centering
\includegraphics{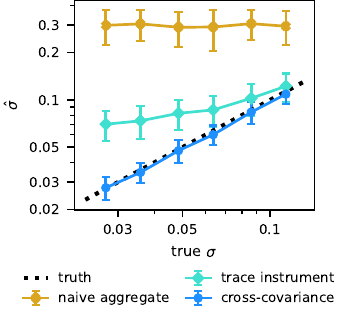}
\caption{The three estimators of $\theta$ for binary classification on the task of
Figure~\ref{fig:sigma}. Average $\hat\sigma$ $\pm$ one standard deviation, over the tasks with
$\that > 0$.}
\label{fig:naive}
\end{figure}

The practical reason is a sign cancellation in the denominator. For a binary softmax head the
output is the logistic function of the logit margin, $\nn(x; \W) = \phi(u(x; \W))$, so at the
pretrained point the probability is $p_0(x) = \phi(u(x; 0))$, and applying the chain rule twice
there yields a Hessian trace proportional to the second derivative of the logistic function,
$t(x) \propto \phi''(u) = p_0(x)\big(1-p_0(x)\big)\big(1-2p_0(x)\big)$
(see the next paragraph). The factor $(1-2p_0)$ flips sign
at $p_0 = \tfrac12$, so on any roughly balanced task the traces nearly cancel in the sum.
On the balanced binary task of Section~\ref{sec:exp-recover} (a logistic model with $400{,}000$
examples and a known true prior) we measure $|\sum_n t_n| / \sum_n |t_n| = 0.003$, meaning the
denominator of \eqref{eq:naive} retains about $0.3$ percent of the available signal, and the
estimate is dominated by noise. It is non-positive in about half of the tasks and otherwise does not
follow the true $\sigma$ (Figure~\ref{fig:naive}).

The second, structural reason applies even without this cancellation. Binary labels satisfy $y^2 = y$, so every
statistic computed from a single example is a function of $\E[p_n]$ alone and informs only the
first moment, which by \eqref{eq:mean} carries only a weak second-order trace signal. The prior
scale controls a variance, so Section~\ref{sec:crossexample_covariances} estimates it from the
covariance between examples.

\paragraph{The trace of a binary softmax head.} Let's write the scalar output as $p = \phi(u)$ with
$u$ the logit margin and $\phi$ the logistic function. Applying the chain rule twice yields the
weight-space Hessian $H = \phi''(u)\,\nabla u\,\nabla u^\top + \phi'(u)\,\nabla^2 u$.
When the adapted weights enter the logit linearly, as for an adapter in the layer producing the
logits, the second term vanishes and $t(x) = \phi''(u)\,\lVert\nabla u\rVert^2$. In
general, the first term, the Gauss-Newton part, carries the same factor. Since
$\phi''(u) = p_0(1-p_0)(1-2p_0)$, the trace inherits the sign flip at $p_0 = \tfrac12$ that
drives the cancellation above.

\paragraph{Regression.} In regression, the residuals are $r_n = y_n - \nn(x_n; 0)$. The
observation noise is independent across examples, so it drops out of all pairs and the noise level
never needs to be known. The per-example second-moment alternative \eqref{eq:regest} can be used when $\tau$ is known and
agrees with the covariance form in our simulations.

\paragraph{Multiclass.} With $C$ classes, the residuals are the vectors $r_n = \mathrm{onehot}(y_n) - p_0(x_n)$, and the gradients
$g_n^{(c)}$ are computed for each class. The pair equations read
$\E[r_{nc}\,r_{mc'}] \approx \theta\,\langle g_n^{(c)}, g_m^{(c')}\rangle$ and range over pairs of
example and class rows. Within one example the multinomial noise is dependent across classes, so
all pairs sharing an example are excluded, which replaces the diagonal subtractions of
\eqref{eq:covest} by block subtractions over the $C\times C$ blocks of each example. On synthetic data, this
estimator recovers the true scale to within a few percent.

\paragraph{Many adapters.} Independence of the per-adapter priors makes cross-adapter response
terms vanish in expectation, which yields \eqref{eq:multimod}. The pooled instance sums the
sketched inner products over adapters before applying \eqref{eq:covest}. The separated instance solves
the linear system $A_\ell = \sum_{\ell'} B_{\ell\ell'}\theta_{\ell'}$ with
$A_\ell$ the residual-weighted sum of inner products of adapter $\ell$ and $B_{\ell\ell'}$ the pairwise
products of inner products, all computable in sketch space. In simulation the separated covariance system
recovers two true scales jointly, and the first-moment analogue fails because the per-adapter
trace profiles correlate at $0.74$ across examples. At real-task sample sizes the separated system
is noise dominated, so the pooled scale is the practical choice.

\paragraph{Token-level generative models.} Next-token prediction is a multiclass problem at every token position. A
practical instance would use the following: the probability of the observed token as the scalar output, one backward pass per
sequence, the residual $1 - p_0(y_t)$ at each position, and only pairs of positions in different
sequences, mirroring the block exclusion above.

\paragraph{A first-moment alternative and its limits.}
\label{app:trace}
Regressing $r_n$ on $\tfrac12 t_n$ by least squares, as \eqref{eq:perpoint} suggests, yields the
\emph{trace-instrument} estimator $\that_{\mathrm{tr}} = 2\sum_n r_nt_n / \sum_n t_n^2$, which is consistent but
label-noise limited, since the trace signal is second order. Its inputs $t_n = \Tr H(x_n)$ can be
estimated with Hutchinson's method \citep{hutchinson1990}, which estimates the trace from a few
Hessian-vector products through the identity $\Tr H = \E[\nu^\top H \nu]$ for random probe vectors
$\nu$ with identity covariance. For adapted linear layers the traces can be computed exactly from the activations, because
the trace reduces to second derivatives with respect to the layer outputs weighted by
squared norms of the layer inputs. Our simulations use the exact form. The variant is useful as a cross-check at
large true scales (Figure~\ref{fig:naive}, where it is non-positive in $27$ to $54\%$ of the tasks and
otherwise overestimates $\sigma$) and is not used in the main pipeline, which needs only first-order
gradients. Per-module variants of \eqref{eq:covest}
exist, but at our sample sizes the module-wise systems are noise dominated, so we report the
pooled scale.

\section{Sketching and its guarantees}
\label{app:sketch}

If $S$ has independent $\mathcal N(0, 1/d)$ entries, then $\E[S^\top S] = I$, and by independence
of the two sides of \eqref{eq:sketch} the sketched inner products are unbiased,
\begin{equation*}
\E\big\langle \tilde g_n,\tilde g_m\big\rangle
= \Tr\big(\E[S_{\mathrm{out}}^\top S_{\mathrm{out}}]\, G_n\, \E[S_{\mathrm{in}}^\top S_{\mathrm{in}}]\,
G_m^\top\big) = \langle g_n, g_m\rangle,
\end{equation*}
including the diagonal entries. Squared inner products acquire a variance bias under a single sketch. Using two
independent sketch pairs and multiplying the inner product from one by that from the other removes it,
since $\E\big[\langle \tilde g^{(1)}_n, \tilde g^{(1)}_m\rangle\langle \tilde g^{(2)}_n, \tilde g^{(2)}_m\rangle\big] = \langle g_n, g_m\rangle^2$. All contractions are
computed in sketch space, for instance $r^\top \tilde G \tilde G^\top r = \lVert \tilde G^\top r\rVert^2$ and
$\langle \tilde G^{(1)} \tilde G^{(1)\top}, \tilde G^{(2)} \tilde G^{(2)\top}\rangle_F = \lVert \tilde G^{(1)\top} \tilde G^{(2)}\rVert_F^2$, with $\tilde G$ the
sketched counterpart of $G$ in Section~\ref{sec:crossexample_covariances}, so no matrix of size $N \times N$ is ever formed.
In a test on a controlled model with sketch width $d = 48$, the estimate of $\theta$ is within about ten percent
of the exact computation, and the scale, which is its fourth root, within three percent.

\section{Posterior approximations, predictive, and evidence}
\label{app:kfac}

The pretrained-model experiments use the Laplace approximation of Laplace-LoRA
\citep{yang2024laplacelora}, computed with its code, which builds on the laplace library
\citep{daxberger2021laplace}. The posterior over the LoRA parameters $w$, the entries of the factors $U$ and $V$ of all
adapted layers, is $\mathcal N\big(w_{\mathrm{MAP}}, (\Lambda + \lambda I)^{-1}\big)$, where
$w_{\mathrm{MAP}}$ are the trained parameters (Section~\ref{sec:exp-mlm}), $\Lambda$ is the
Kronecker-factored generalized Gauss-Newton matrix of the training loss at $w_{\mathrm{MAP}}$ and
$\lambda$ is the prior precision. Predictions use the linearized predictive
\citep{immer2021improving}: the logits at $x$ are Gaussian, centred at those of $w_{\mathrm{MAP}}$,
with covariance $J(x)(\Lambda + \lambda I)^{-1}J(x)^\top$, where $J(x)$ is the Jacobian of the logits
with respect to $w$, and the class probabilities are the softmax averaged over samples of the
logits. The mutual information of Section~\ref{sec:exp-ood} is the entropy of this averaged
prediction minus the average entropy of the sampled predictions.

Marginal likelihood chooses $\lambda$ by maximizing the Laplace approximation of the evidence,
\begin{equation*}
\log p(\mathcal D \mid w_{\mathrm{MAP}}) - \tfrac12\Big(\log\det(\Lambda + \lambda I) - \dim(w)\log\lambda
+ \lambda\lVert w_{\mathrm{MAP}}\rVert^2\Big),
\end{equation*}
with $\dim(w)$ the number of LoRA parameters, by gradient steps on $\log\lambda$. Validation search performs
gradient steps on $\log\lambda$ that lower the NLL of the predictive on the validation set, starting
from the marginal-likelihood value. The MAP uses $\lambda = 10^8$ in the same predictive. A
temperature $T$ replaces the predictive probabilities $p$ by $p^{1/T}$, renormalized, and is fitted on
the validation set. Appendix~\ref{app:mlm} reports the numbers of steps, samples and temperatures.

\section{Toy experiments}
\label{app:toy}

\paragraph{Second-order moments (Section~\ref{sec:exp-valid}).} The network is a two-layer tanh
classifier with 20 inputs and 12 hidden units, with the update on the first layer. The moments are
estimated at six inputs from $40{,}000$ draws per $K$ and $\theta\bar s$.

\paragraph{Prior scale (Section~\ref{sec:exp-recover}).} For regression the network is
$x \mapsto w^\top \tanh\big((W_0 + \W)\, x\big)$ with 32 inputs and 16 hidden units. Each task has $N = 4{,}000$ examples
and label noise $\tau = \tfrac12 (\theta\bar s)^{1/2}$, which keeps the ratio of signal to noise
fixed across $\sigma$. For binary classification the network is a logistic model with 32 inputs, and
since the Bernoulli noise does not shrink with $\sigma$, each task has $N = 400{,}000$ examples. There are 80 tasks per scale
for regression and 96 for classification.

\paragraph{Rank (Section~\ref{sec:exp-K}).} The networks are $x \mapsto \tanh\big((W_0 + \W)\, x\big)$
with 112 inputs and 112 hidden units, followed by a fixed output layer. Each of the 24 examples averages the
outputs at two inputs, so that $\rho_n$ ranges from $0.67$ to $0.90$. The study is repeated 10 times
for each $K$ from 10 to 100 and each $\theta\bar s \in \{2\cdot10^{-4}, 2\cdot10^{-3}, 10^{-2}\}$.

\section{Rank transfer in fine-tuning}
\label{app:rank}

This appendix presents the protocol and the full results of the rank study in Section~\ref{sec:exp-rank}.
The model, the training and the posterior are those of Appendix~\ref{app:mlm}
with $1{,}024$ examples, $819$ for training and $205$ for validation, except that the LoRA rank is
$r \in \{4, 8, 16, 32, 64\}$. The tasks are AG News and CR. For each run we evaluate the validation
and test NLL of the predictive at 41 precisions from $10^{-2}$ to $10^{6}$, equally spaced on a log
scale, and define $\lambda^\ast(r)$ as the minimizer of the validation NLL. In every run it lies inside
this range. The seeds are 0, 1 and 2 for AG News and 1, 2 and 3 for CR. For CR with seed 0 the
evaluation at the 41 precisions failed at two ranks, because the covariance of the logits was
numerically not positive definite, and we replaced this seed. We fit the slope of $\log\lambda$
against $\log r$ by least squares with one intercept per task, with a $95\%$ bootstrap interval over
the 30 runs.

Each task alone yields the same slope of $\lambda^\ast$ as the two tasks together, close to $-1$, and
so do the precisions chosen by validation search and by marginal likelihood.
Table~\ref{tab:rank-large} reports $\lambda^\ast$, $\lhat$ and the test NLL at every rank. On AG News
one seed yields a much smaller $\that$, which explains the large spread of our NLL between seeds. In
Figure~\ref{fig:rank-transfer}, the validation NLL at $\lhat$ is interpolated between the two nearest of
the 41 precisions.

Appendix~\ref{app:linearized} explains the slope of $-1$ by the variance of the linearized predictive,
which scales as $1/(r\lambda)$. The runs confirm this scaling. Let $\omega$ be the variance of a logit under
this predictive, averaged over the test examples and classes. Within a run, $\omega$ is inversely
proportional to the precision. For example, ours and validation search choose precisions up to 30
times apart, yet their products $\lambda\omega$ agree within $10\%$. Across ranks, $r\lambda\omega$ shows no
trend.

\begin{table}[h]
\centering\footnotesize
\caption{Rank study on RoBERTa-large with $1{,}024$ training examples. $\lambda^\ast$ and $\lhat$:
median over three seeds. Test NLL: mean over the seeds $\pm$ their standard deviation, at
$\lambda^\ast$ and for each method. ``+ T'' adds a temperature fitted on the validation set.}
\label{tab:rank-large}
\setlength{\tabcolsep}{5pt}
\begin{tabular}{lrrr ccccc}
\toprule
 & & & & \multicolumn{5}{c}{test NLL} \\
\cmidrule(lr){5-9}
 & & & & & \multicolumn{2}{c}{ours} & \multicolumn{2}{c}{validation search} \\
\cmidrule(lr){6-7}\cmidrule(lr){8-9}
task & $r$ & $\lambda^\ast$ & $\lhat$ & at $\lambda^\ast$ & & + T & & + T \\
\midrule
AG News & 4 & 2{,}512 & 3{,}340 & 0.308 $\pm$ 0.009 & 0.367 $\pm$ 0.113 & 0.302 $\pm$ 0.003 & 0.306 $\pm$ 0.003 & 0.305 $\pm$ 0.005 \\
 & 8 & 1{,}585 & 2{,}362 & 0.304 $\pm$ 0.009 & 0.380 $\pm$ 0.129 & 0.303 $\pm$ 0.011 & 0.304 $\pm$ 0.014 & 0.301 $\pm$ 0.005 \\
 & 16 & 631 & 1{,}670 & 0.301 $\pm$ 0.005 & 0.411 $\pm$ 0.128 & 0.300 $\pm$ 0.008 & 0.301 $\pm$ 0.006 & 0.301 $\pm$ 0.007 \\
 & 32 & 398 & 1{,}181 & 0.306 $\pm$ 0.007 & 0.448 $\pm$ 0.134 & 0.305 $\pm$ 0.006 & 0.305 $\pm$ 0.007 & 0.303 $\pm$ 0.005 \\
 & 64 & 158 & 835 & 0.300 $\pm$ 0.006 & 0.486 $\pm$ 0.146 & 0.304 $\pm$ 0.007 & 0.302 $\pm$ 0.005 & 0.301 $\pm$ 0.004 \\
\midrule
CR & 4 & 631 & 988 & 0.189 $\pm$ 0.021 & 0.193 $\pm$ 0.020 & 0.190 $\pm$ 0.017 & 0.197 $\pm$ 0.024 & 0.188 $\pm$ 0.016 \\
 & 8 & 398 & 699 & 0.195 $\pm$ 0.042 & 0.210 $\pm$ 0.043 & 0.191 $\pm$ 0.029 & 0.191 $\pm$ 0.022 & 0.192 $\pm$ 0.032 \\
 & 16 & 158 & 494 & 0.193 $\pm$ 0.025 & 0.231 $\pm$ 0.045 & 0.194 $\pm$ 0.034 & 0.197 $\pm$ 0.043 & 0.193 $\pm$ 0.033 \\
 & 32 & 63 & 349 & 0.195 $\pm$ 0.034 & 0.256 $\pm$ 0.053 & 0.197 $\pm$ 0.039 & 0.200 $\pm$ 0.041 & 0.200 $\pm$ 0.041 \\
 & 64 & 63 & 247 & 0.187 $\pm$ 0.020 & 0.279 $\pm$ 0.049 & 0.187 $\pm$ 0.016 & 0.191 $\pm$ 0.014 & 0.191 $\pm$ 0.021 \\
\bottomrule

\end{tabular}
\end{table}

\section{Rank scaling of the prior and of the linearized predictive}
\label{app:linearized}

Corollary~\ref{cor:rank} holds for the prior, and the measurement of Section~\ref{sec:exp-rank} does
not contradict it. The two yield different rank scalings because they concern different predictives.
The corollary keeps the prior predictive fixed, and under the prior the variance of $\W$ is quadratic
in $1/\lambda$. After fine-tuning, the Laplace posterior is used through a predictive linearized at
the trained factors. This is the standard practical choice, because the linearized predictive predicts
better than a predictive that evaluates the network at samples of the weights \citep{immer2021improving}. Its variance
is linear in $1/\lambda$ and, at a fixed $\lambda$, falls as $1/r$. The difference does not come from the range of the
second-order expansion of Section~\ref{sec:propagation}. It appears even for a network that is
linear in $\W$, because it comes from the factorization of $\W$.

Let's write the factors as their trained values plus a perturbation, $U = U_{\mathrm{MAP}} + \Delta U$ and
$V = V_{\mathrm{MAP}} + \Delta V$. Then
\begin{equation}
\W = \frac{\alpha}{r}\,U_{\mathrm{MAP}} V_{\mathrm{MAP}}
+ \frac{\alpha}{r}\big(\Delta U\, V_{\mathrm{MAP}} + U_{\mathrm{MAP}}\, \Delta V\big)
+ \frac{\alpha}{r}\,\Delta U\, \Delta V .
\label{eq:factor-split}
\end{equation}
Under the prior the factors are centred at zero, so, with $U_{\mathrm{MAP}}$ and $V_{\mathrm{MAP}}$ replaced by zero, only the last term is
present. By \eqref{eq:lambda} its entries have variance $\alpha^2/(r\lambda^2)$, and keeping this
variance fixed yields $\lambda \propto r^{-1/2}$.

The Laplace posterior of Appendix~\ref{app:kfac} is centred at the trained factors. Its covariance
$(\Lambda + \lambda I)^{-1}$ is close to $\lambda^{-1} I$ in the directions where $\lambda$ dominates
$\Lambda$, which with few training examples and many LoRA parameters are most directions. The
linearized predictive replaces the network by its first-order expansion in the LoRA parameters at
their trained values. It therefore retains the middle term of \eqref{eq:factor-split} and omits the last
one. If the entries of $\Delta U$ and $\Delta V$ are independent with variance $1/\lambda$, the entry
$(i,j)$ of the middle term has variance
\begin{equation}
\frac{\alpha^2}{r^2\lambda}\Big(\sum_{k=1}^{r} V_{\mathrm{MAP},kj}^2 + \sum_{k=1}^{r} U_{\mathrm{MAP},ik}^2\Big),
\label{eq:linvar}
\end{equation}
which is linear in $1/\lambda$. When the entries of the trained factors have a size that does not
depend on the rank, each sum grows as $r$, and \eqref{eq:linvar} scales as $\alpha^2/(r\lambda)$. In
LoRA, $V$ is initialized with a scale set by the input dimension and $U$ at zero, so this holds as
long as fine-tuning does not change the scale of the factors with the rank. The runs of
Appendix~\ref{app:rank} confirm the resulting scaling for the logits. The precision that keeps the
variance of the linearized predictive at the level that predicts best then falls as $1/r$, the slope
measured in Section~\ref{sec:exp-rank}.

Our estimate sets the prior, and $\lhat \propto r^{-1/2}$ is the rank scaling of the prior. Used
directly as the precision of the linearized predictive, it moves away from the best precision by a
factor that grows as $\sqrt{r}$, and a temperature fitted on validation data compensates for this
(Section~\ref{sec:exp-rank}). A rank scaling for the precision of the linearized predictive would have
to account for the trained factors through \eqref{eq:linvar}, and they are not available before
fine-tuning.

A predictive that samples the factors from the posterior and evaluates the network at each sample,
instead of linearizing it, retains the last term of \eqref{eq:factor-split}. Its variance is then the
sum of \eqref{eq:linvar} and $\alpha^2/(r\lambda^2)$, which scales with the rank as under the prior.
The best precision for this predictive would fall with the rank between $r^{-1}$ and $r^{-1/2}$,
depending on which of the two parts dominates. We have not evaluated this predictive.

\clearpage
\section{RoBERTa-large experiments}
\label{app:mlm}

\paragraph{Data.} The test sets are the GLUE validation set of SST-2
\citep{socher2013recursive,wang2019glue} ($872$ examples) and the test sets of MR
\citep{pang2005seeing} ($1{,}066$), CR \citep{hu2004mining} ($753$) and AG News
\citep{zhang2015character} ($7{,}600$), from the Hugging
Face hub (\texttt{cornell-movie-review-data/rotten\_tomatoes}, \texttt{SetFit/SentEval-CR},
\texttt{fancyzhx/ag\_news}). For each seed the training split is shuffled and its first $n$
examples are kept (for CR at $n = 4{,}096$, all $3{,}012$ examples). The first $20\%$ of them, rounded up, form
the validation set and the rest the training set. 

\paragraph{Classifier and training.} Each input $x$ is placed in a template with one masked
position, ``$x$ It was \texttt{<mask>} .'' with the label words \emph{terrible} and
\emph{great} for the sentiment tasks, and ``$x$ ( \texttt{<mask>} )'' with \emph{World},
\emph{Sports}, \emph{Business} and \emph{Tech} for AG News. The class logits are the scores of the
frozen masked-language-model head for these words at the masked position. LoRA with $r = 8$,
$\alpha = 16$ and dropout $0.1$ adapts the query and value projections of all 24 layers,
with $786{,}432$ parameters in total. Training uses Adam without weight decay, a batch size of 32, inputs of at most
256 tokens, and $10{,}000$ steps with a learning rate of $5\cdot10^{-5}$ decayed linearly to zero.

\paragraph{The estimate.} $\lhat = \alpha/\sqrt{\that r}$ is computed at $\W = 0$ from the
training set by Algorithm~\ref{alg:main}, with sketches of size $d = 32$.

\paragraph{Posterior, choices of $\lambda$ and metrics.} The Laplace approximation of
Appendix~\ref{app:kfac} covers all LoRA parameters $w$, with the
Kronecker-factored generalized Gauss-Newton matrix $\Lambda$, as implemented in the code of Laplace-LoRA
\citep{yang2024laplacelora}. Predictions average the softmax over $100{,}000$
samples of the logits from the linearized predictive, with the full covariance between classes.
Marginal likelihood is maximized by 100 gradient steps with a learning rate of $0.1$. Validation search
performs 100 gradient steps with a learning rate of $0.1$ on the validation NLL, starting from the
marginal-likelihood $\lambda$. The MAP uses $\lambda = 10^{8}$ in the same predictive. A temperature
is chosen from 81 values between $0.1$ and $10$, equally spaced on a log scale, by the validation
NLL. ECE uses 10 bins of equal width in confidence.

\begin{table}[tbp]
\centering\footnotesize\renewcommand{\arraystretch}{0.85}
\caption{Results on AG News: mean over three seeds $\pm$ their standard deviation. Zero-shot does
not depend on the seed. Bold: the best value in each column of a block, among the methods without
validation data (upper rows) and among those with it (lower rows).}
\label{tab:mlm-agnews}
\begin{tabular}{l cccc}
\toprule
 & \multicolumn{4}{c}{training examples} \\
 & 64 & 256 & $1{,}024$ & $4{,}096$ \\
\midrule
\emph{test NLL} & & & & \\
zero-shot & 0.746 & 0.746 & 0.746 & 0.746 \\
MAP & 1.007 $\pm$ 0.149 & 0.875 $\pm$ 0.136 & 0.747 $\pm$ 0.035 & 0.473 $\pm$ 0.039 \\
marginal likelihood & 1.336 $\pm$ 0.018 & 1.250 $\pm$ 0.019 & 1.034 $\pm$ 0.021 & 0.354 $\pm$ 0.019 \\
ours & \textbf{0.458} $\pm$ 0.054 & \textbf{0.423} $\pm$ 0.109 & \textbf{0.395} $\pm$ 0.147 & \textbf{0.277} $\pm$ 0.012 \\
\cmidrule(lr){2-5}
MAP + T & \textbf{0.439} $\pm$ 0.072 & \textbf{0.363} $\pm$ 0.020 & 0.314 $\pm$ 0.006 & 0.259 $\pm$ 0.012 \\
marginal likelihood + T & 1.394 $\pm$ 0.164 & 0.937 $\pm$ 0.093 & 0.429 $\pm$ 0.052 & \textbf{0.255} $\pm$ 0.007 \\
ours + T & 0.442 $\pm$ 0.066 & 0.364 $\pm$ 0.016 & \textbf{0.307} $\pm$ 0.006 & 0.259 $\pm$ 0.008 \\
validation search & \textbf{0.439} $\pm$ 0.070 & 0.367 $\pm$ 0.010 & \textbf{0.307} $\pm$ 0.008 & 0.258 $\pm$ 0.008 \\
\midrule
\emph{test ECE} & & & & \\
zero-shot & \textbf{0.034} & \textbf{0.034} & \textbf{0.034} & 0.034 \\
MAP & 0.119 $\pm$ 0.018 & 0.095 $\pm$ 0.007 & 0.084 $\pm$ 0.002 & 0.064 $\pm$ 0.007 \\
marginal likelihood & 0.106 $\pm$ 0.065 & 0.311 $\pm$ 0.068 & 0.474 $\pm$ 0.012 & 0.145 $\pm$ 0.011 \\
ours & 0.038 $\pm$ 0.016 & 0.039 $\pm$ 0.021 & \textbf{0.034} $\pm$ 0.028 & \textbf{0.022} $\pm$ 0.002 \\
\cmidrule(lr){2-5}
MAP + T & \textbf{0.037} $\pm$ 0.041 & \textbf{0.021} $\pm$ 0.008 & \textbf{0.015} $\pm$ 0.005 & \textbf{0.008} $\pm$ 0.002 \\
marginal likelihood + T & 0.134 $\pm$ 0.094 & 0.083 $\pm$ 0.030 & 0.027 $\pm$ 0.035 & 0.013 $\pm$ 0.004 \\
ours + T & 0.073 $\pm$ 0.042 & 0.041 $\pm$ 0.021 & 0.030 $\pm$ 0.014 & 0.018 $\pm$ 0.004 \\
validation search & 0.071 $\pm$ 0.045 & 0.044 $\pm$ 0.029 & 0.030 $\pm$ 0.012 & 0.018 $\pm$ 0.003 \\
\midrule
\emph{test accuracy} & & & & \\
zero-shot & 0.700 & 0.700 & 0.700 & 0.700 \\
MAP & \textbf{0.855} $\pm$ 0.021 & \textbf{0.887} $\pm$ 0.005 & \textbf{0.902} $\pm$ 0.002 & \textbf{0.916} $\pm$ 0.006 \\
marginal likelihood & 0.398 $\pm$ 0.065 & 0.615 $\pm$ 0.071 & 0.842 $\pm$ 0.011 & 0.915 $\pm$ 0.005 \\
ours & \textbf{0.855} $\pm$ 0.021 & \textbf{0.887} $\pm$ 0.005 & \textbf{0.902} $\pm$ 0.002 & \textbf{0.916} $\pm$ 0.005 \\
\cmidrule(lr){2-5}
MAP + T & \textbf{0.855} $\pm$ 0.021 & \textbf{0.887} $\pm$ 0.005 & \textbf{0.902} $\pm$ 0.002 & \textbf{0.916} $\pm$ 0.006 \\
marginal likelihood + T & 0.398 $\pm$ 0.065 & 0.615 $\pm$ 0.071 & 0.842 $\pm$ 0.011 & 0.915 $\pm$ 0.005 \\
ours + T & \textbf{0.855} $\pm$ 0.021 & \textbf{0.887} $\pm$ 0.005 & \textbf{0.902} $\pm$ 0.002 & \textbf{0.916} $\pm$ 0.005 \\
validation search & 0.854 $\pm$ 0.022 & \textbf{0.887} $\pm$ 0.005 & \textbf{0.902} $\pm$ 0.002 & \textbf{0.916} $\pm$ 0.005 \\
\bottomrule

\end{tabular}
\par\medskip
\caption{Results on CR, as in Table~\ref{tab:mlm-agnews}. Two seeds at 256 examples.}
\label{tab:mlm-cr}
\begin{tabular}{l cccc}
\toprule
 & \multicolumn{4}{c}{training examples} \\
 & 64 & 256 & $1{,}024$ & $4{,}096$ \\
\midrule
\emph{test NLL} & & & & \\
zero-shot & 0.448 & 0.448 & 0.448 & 0.448 \\
MAP & 0.688 $\pm$ 0.017 & 0.601 $\pm$ 0.139 & 0.574 $\pm$ 0.054 & 0.440 $\pm$ 0.016 \\
marginal likelihood & 0.633 $\pm$ 0.011 & 0.552 $\pm$ 0.005 & 0.475 $\pm$ 0.018 & 0.322 $\pm$ 0.009 \\
ours & \textbf{0.326} $\pm$ 0.049 & \textbf{0.322} $\pm$ 0.058 & \textbf{0.218} $\pm$ 0.034 & \textbf{0.170} $\pm$ 0.013 \\
\cmidrule(lr){2-5}
MAP + T & \textbf{0.225} $\pm$ 0.015 & 0.218 $\pm$ 0.016 & 0.200 $\pm$ 0.020 & 0.173 $\pm$ 0.002 \\
marginal likelihood + T & 0.312 $\pm$ 0.036 & \textbf{0.209} $\pm$ 0.007 & 0.189 $\pm$ 0.025 & \textbf{0.160} $\pm$ 0.007 \\
ours + T & 0.228 $\pm$ 0.031 & 0.214 $\pm$ 0.012 & \textbf{0.188} $\pm$ 0.024 & 0.164 $\pm$ 0.008 \\
validation search & 0.235 $\pm$ 0.027 & 0.215 $\pm$ 0.013 & 0.190 $\pm$ 0.028 & 0.165 $\pm$ 0.012 \\
\midrule
\emph{test ECE} & & & & \\
zero-shot & 0.076 & 0.076 & 0.076 & 0.076 \\
MAP & 0.072 $\pm$ 0.004 & 0.067 $\pm$ 0.009 & 0.060 $\pm$ 0.008 & 0.054 $\pm$ 0.003 \\
marginal likelihood & 0.390 $\pm$ 0.006 & 0.343 $\pm$ 0.009 & 0.303 $\pm$ 0.015 & 0.191 $\pm$ 0.007 \\
ours & \textbf{0.042} $\pm$ 0.004 & \textbf{0.040} $\pm$ 0.018 & \textbf{0.035} $\pm$ 0.006 & \textbf{0.023} $\pm$ 0.006 \\
\cmidrule(lr){2-5}
MAP + T & \textbf{0.037} $\pm$ 0.019 & 0.029 $\pm$ 0.021 & 0.024 $\pm$ 0.015 & \textbf{0.014} $\pm$ 0.001 \\
marginal likelihood + T & 0.150 $\pm$ 0.036 & \textbf{0.027} $\pm$ 0.016 & \textbf{0.022} $\pm$ 0.011 & 0.021 $\pm$ 0.003 \\
ours + T & 0.053 $\pm$ 0.039 & 0.035 $\pm$ 0.013 & \textbf{0.022} $\pm$ 0.006 & 0.023 $\pm$ 0.002 \\
validation search & 0.052 $\pm$ 0.038 & 0.037 $\pm$ 0.012 & 0.023 $\pm$ 0.006 & 0.022 $\pm$ 0.003 \\
\midrule
\emph{test accuracy} & & & & \\
zero-shot & 0.810 & 0.810 & 0.810 & 0.810 \\
MAP & 0.923 $\pm$ 0.003 & \textbf{0.928} $\pm$ 0.008 & \textbf{0.936} $\pm$ 0.009 & \textbf{0.940} $\pm$ 0.004 \\
marginal likelihood & 0.923 $\pm$ 0.001 & 0.926 $\pm$ 0.005 & \textbf{0.936} $\pm$ 0.008 & \textbf{0.940} $\pm$ 0.003 \\
ours & \textbf{0.924} $\pm$ 0.003 & \textbf{0.928} $\pm$ 0.008 & \textbf{0.936} $\pm$ 0.009 & \textbf{0.940} $\pm$ 0.004 \\
\cmidrule(lr){2-5}
MAP + T & 0.923 $\pm$ 0.003 & \textbf{0.928} $\pm$ 0.008 & \textbf{0.936} $\pm$ 0.009 & \textbf{0.940} $\pm$ 0.004 \\
marginal likelihood + T & 0.923 $\pm$ 0.001 & 0.926 $\pm$ 0.005 & \textbf{0.936} $\pm$ 0.008 & \textbf{0.940} $\pm$ 0.003 \\
ours + T & \textbf{0.924} $\pm$ 0.003 & \textbf{0.928} $\pm$ 0.008 & \textbf{0.936} $\pm$ 0.009 & \textbf{0.940} $\pm$ 0.004 \\
validation search & 0.923 $\pm$ 0.003 & \textbf{0.928} $\pm$ 0.008 & \textbf{0.936} $\pm$ 0.008 & \textbf{0.940} $\pm$ 0.004 \\
\bottomrule

\end{tabular}
\end{table}

\begin{table}[tbp]
\centering\footnotesize\renewcommand{\arraystretch}{0.85}
\caption{Results on MR, as in Table~\ref{tab:mlm-agnews}.}
\label{tab:mlm-mr}
\begin{tabular}{l cccc}
\toprule
 & \multicolumn{4}{c}{training examples} \\
 & 64 & 256 & $1{,}024$ & $4{,}096$ \\
\midrule
\emph{test NLL} & & & & \\
zero-shot & \textbf{0.461} & 0.461 & 0.461 & 0.461 \\
MAP & 0.936 $\pm$ 0.055 & 0.895 $\pm$ 0.062 & 0.807 $\pm$ 0.060 & 0.635 $\pm$ 0.048 \\
marginal likelihood & 0.644 $\pm$ 0.015 & 0.593 $\pm$ 0.009 & 0.514 $\pm$ 0.011 & \textbf{0.284} $\pm$ 0.004 \\
ours & 0.477 $\pm$ 0.069 & \textbf{0.416} $\pm$ 0.020 & \textbf{0.331} $\pm$ 0.020 & 0.296 $\pm$ 0.006 \\
\cmidrule(lr){2-5}
MAP + T & 1.303 $\pm$ 0.871 & 0.288 $\pm$ 0.020 & 0.264 $\pm$ 0.003 & \textbf{0.251} $\pm$ 0.009 \\
marginal likelihood + T & \textbf{0.385} $\pm$ 0.064 & \textbf{0.270} $\pm$ 0.007 & \textbf{0.262} $\pm$ 0.006 & 0.253 $\pm$ 0.005 \\
ours + T & 0.994 $\pm$ 0.587 & 0.312 $\pm$ 0.092 & 0.266 $\pm$ 0.005 & 0.257 $\pm$ 0.005 \\
validation search & 0.493 $\pm$ 0.166 & 0.310 $\pm$ 0.090 & 0.263 $\pm$ 0.006 & 0.253 $\pm$ 0.003 \\
\midrule
\emph{test ECE} & & & & \\
zero-shot & \textbf{0.035} & \textbf{0.035} & \textbf{0.035} & \textbf{0.035} \\
MAP & 0.105 $\pm$ 0.008 & 0.093 $\pm$ 0.002 & 0.085 $\pm$ 0.002 & 0.079 $\pm$ 0.007 \\
marginal likelihood & 0.353 $\pm$ 0.002 & 0.341 $\pm$ 0.009 & 0.293 $\pm$ 0.009 & 0.089 $\pm$ 0.013 \\
ours & 0.057 $\pm$ 0.009 & 0.050 $\pm$ 0.005 & 0.039 $\pm$ 0.004 & 0.038 $\pm$ 0.006 \\
\cmidrule(lr){2-5}
MAP + T & 0.091 $\pm$ 0.047 & 0.042 $\pm$ 0.009 & \textbf{0.016} $\pm$ 0.004 & \textbf{0.022} $\pm$ 0.001 \\
marginal likelihood + T & 0.148 $\pm$ 0.050 & 0.043 $\pm$ 0.008 & 0.030 $\pm$ 0.007 & 0.027 $\pm$ 0.002 \\
ours + T & 0.090 $\pm$ 0.039 & 0.033 $\pm$ 0.013 & 0.035 $\pm$ 0.008 & 0.031 $\pm$ 0.003 \\
validation search & \textbf{0.067} $\pm$ 0.013 & \textbf{0.030} $\pm$ 0.017 & 0.031 $\pm$ 0.007 & 0.030 $\pm$ 0.004 \\
\midrule
\emph{test accuracy} & & & & \\
zero-shot & 0.786 & 0.786 & 0.786 & 0.786 \\
MAP & \textbf{0.882} $\pm$ 0.013 & \textbf{0.900} $\pm$ 0.006 & 0.906 $\pm$ 0.002 & \textbf{0.910} $\pm$ 0.006 \\
marginal likelihood & 0.881 $\pm$ 0.011 & \textbf{0.900} $\pm$ 0.007 & \textbf{0.907} $\pm$ 0.001 & \textbf{0.910} $\pm$ 0.007 \\
ours & \textbf{0.882} $\pm$ 0.013 & \textbf{0.900} $\pm$ 0.006 & 0.906 $\pm$ 0.002 & \textbf{0.910} $\pm$ 0.006 \\
\cmidrule(lr){2-5}
MAP + T & \textbf{0.882} $\pm$ 0.013 & \textbf{0.900} $\pm$ 0.006 & 0.906 $\pm$ 0.002 & \textbf{0.910} $\pm$ 0.006 \\
marginal likelihood + T & 0.881 $\pm$ 0.011 & \textbf{0.900} $\pm$ 0.007 & \textbf{0.907} $\pm$ 0.001 & \textbf{0.910} $\pm$ 0.007 \\
ours + T & \textbf{0.882} $\pm$ 0.013 & \textbf{0.900} $\pm$ 0.006 & 0.906 $\pm$ 0.002 & \textbf{0.910} $\pm$ 0.006 \\
validation search & \textbf{0.882} $\pm$ 0.013 & \textbf{0.900} $\pm$ 0.006 & 0.906 $\pm$ 0.002 & \textbf{0.910} $\pm$ 0.006 \\
\bottomrule

\end{tabular}
\par\medskip
\caption{Results on SST-2, as in Table~\ref{tab:mlm-agnews}. Two seeds at $4{,}096$ examples.}
\label{tab:mlm-sst2}
\begin{tabular}{l cccc}
\toprule
 & \multicolumn{4}{c}{training examples} \\
 & 64 & 256 & $1{,}024$ & $4{,}096$ \\
\midrule
\emph{test NLL} & & & & \\
zero-shot & 0.384 & 0.384 & 0.384 & 0.384 \\
MAP & 0.597 $\pm$ 0.085 & 0.512 $\pm$ 0.116 & 0.479 $\pm$ 0.019 & 0.332 $\pm$ 0.024 \\
marginal likelihood & 0.621 $\pm$ 0.015 & 0.540 $\pm$ 0.027 & 0.412 $\pm$ 0.015 & \textbf{0.160} $\pm$ 0.000 \\
ours & \textbf{0.326} $\pm$ 0.057 & \textbf{0.291} $\pm$ 0.060 & \textbf{0.231} $\pm$ 0.012 & 0.203 $\pm$ 0.022 \\
\cmidrule(lr){2-5}
MAP + T & 0.539 $\pm$ 0.466 & 0.208 $\pm$ 0.018 & 0.176 $\pm$ 0.007 & 0.156 $\pm$ 0.007 \\
marginal likelihood + T & 0.285 $\pm$ 0.049 & 0.204 $\pm$ 0.017 & \textbf{0.164} $\pm$ 0.005 & 0.151 $\pm$ 0.008 \\
ours + T & 0.414 $\pm$ 0.296 & \textbf{0.202} $\pm$ 0.021 & 0.165 $\pm$ 0.005 & \textbf{0.150} $\pm$ 0.007 \\
validation search & \textbf{0.274} $\pm$ 0.055 & 0.207 $\pm$ 0.021 & 0.167 $\pm$ 0.004 & 0.152 $\pm$ 0.008 \\
\midrule
\emph{test ECE} & & & & \\
zero-shot & \textbf{0.018} & \textbf{0.018} & \textbf{0.018} & \textbf{0.018} \\
MAP & 0.075 $\pm$ 0.009 & 0.066 $\pm$ 0.012 & 0.055 $\pm$ 0.002 & 0.043 $\pm$ 0.003 \\
marginal likelihood & 0.373 $\pm$ 0.010 & 0.336 $\pm$ 0.011 & 0.263 $\pm$ 0.014 & 0.037 $\pm$ 0.009 \\
ours & 0.041 $\pm$ 0.007 & 0.040 $\pm$ 0.010 & 0.036 $\pm$ 0.003 & 0.026 $\pm$ 0.004 \\
\cmidrule(lr){2-5}
MAP + T & 0.079 $\pm$ 0.011 & 0.034 $\pm$ 0.016 & \textbf{0.020} $\pm$ 0.013 & 0.014 $\pm$ 0.003 \\
marginal likelihood + T & 0.113 $\pm$ 0.037 & 0.040 $\pm$ 0.015 & 0.023 $\pm$ 0.005 & 0.013 $\pm$ 0.004 \\
ours + T & 0.057 $\pm$ 0.013 & \textbf{0.029} $\pm$ 0.021 & 0.023 $\pm$ 0.006 & \textbf{0.012} $\pm$ 0.003 \\
validation search & \textbf{0.048} $\pm$ 0.010 & 0.031 $\pm$ 0.023 & 0.024 $\pm$ 0.006 & 0.014 $\pm$ 0.006 \\
\midrule
\emph{test accuracy} & & & & \\
zero-shot & 0.817 & 0.817 & 0.817 & 0.817 \\
MAP & \textbf{0.915} $\pm$ 0.010 & 0.924 $\pm$ 0.012 & \textbf{0.940} $\pm$ 0.004 & \textbf{0.947} $\pm$ 0.003 \\
marginal likelihood & 0.913 $\pm$ 0.015 & \textbf{0.925} $\pm$ 0.009 & \textbf{0.940} $\pm$ 0.004 & \textbf{0.947} $\pm$ 0.002 \\
ours & 0.914 $\pm$ 0.010 & 0.924 $\pm$ 0.012 & \textbf{0.940} $\pm$ 0.004 & \textbf{0.947} $\pm$ 0.003 \\
\cmidrule(lr){2-5}
MAP + T & \textbf{0.915} $\pm$ 0.010 & 0.924 $\pm$ 0.012 & \textbf{0.940} $\pm$ 0.004 & \textbf{0.947} $\pm$ 0.003 \\
marginal likelihood + T & 0.913 $\pm$ 0.015 & \textbf{0.925} $\pm$ 0.009 & \textbf{0.940} $\pm$ 0.004 & \textbf{0.947} $\pm$ 0.002 \\
ours + T & 0.914 $\pm$ 0.010 & 0.924 $\pm$ 0.012 & \textbf{0.940} $\pm$ 0.004 & \textbf{0.947} $\pm$ 0.003 \\
validation search & 0.914 $\pm$ 0.010 & 0.924 $\pm$ 0.012 & \textbf{0.940} $\pm$ 0.004 & \textbf{0.947} $\pm$ 0.002 \\
\bottomrule

\end{tabular}
\end{table}

\begin{table}[tbp]
\centering\small
\caption{AUROC for separating OOD inputs from each task's test inputs, with the mutual
information as the score: mean over three seeds $\pm$ their standard deviation (two seeds for CR at 256 and SST-2
at $4{,}096$ examples). Bold: the best value in each column of a block.}
\label{tab:mlm-ood}
\begin{tabular}{l cccc}
\toprule
 & \multicolumn{4}{c}{training examples} \\
 & 64 & 256 & $1{,}024$ & $4{,}096$ \\
\midrule
\emph{AG News, OOD inputs from SST-2} & & & & \\
marginal likelihood & 0.360 $\pm$ 0.168 & 0.396 $\pm$ 0.038 & 0.745 $\pm$ 0.063 & \textbf{0.946} $\pm$ 0.006 \\
ours & 0.861 $\pm$ 0.017 & 0.829 $\pm$ 0.077 & 0.871 $\pm$ 0.051 & 0.904 $\pm$ 0.002 \\
validation search & \textbf{0.885} $\pm$ 0.024 & \textbf{0.838} $\pm$ 0.092 & \textbf{0.892} $\pm$ 0.040 & 0.922 $\pm$ 0.005 \\
\midrule
\emph{CR, OOD inputs from AG News} & & & & \\
marginal likelihood & 0.697 $\pm$ 0.052 & 0.781 $\pm$ 0.028 & 0.768 $\pm$ 0.011 & 0.772 $\pm$ 0.016 \\
ours & \textbf{0.839} $\pm$ 0.029 & \textbf{0.851} $\pm$ 0.030 & \textbf{0.838} $\pm$ 0.009 & \textbf{0.800} $\pm$ 0.019 \\
validation search & 0.819 $\pm$ 0.041 & 0.833 $\pm$ 0.035 & 0.827 $\pm$ 0.011 & 0.795 $\pm$ 0.016 \\
\midrule
\emph{MR, OOD inputs from AG News} & & & & \\
marginal likelihood & 0.503 $\pm$ 0.204 & 0.655 $\pm$ 0.053 & 0.667 $\pm$ 0.020 & 0.749 $\pm$ 0.018 \\
ours & \textbf{0.744} $\pm$ 0.086 & \textbf{0.796} $\pm$ 0.038 & \textbf{0.780} $\pm$ 0.036 & \textbf{0.772} $\pm$ 0.014 \\
validation search & \textbf{0.744} $\pm$ 0.075 & 0.779 $\pm$ 0.029 & 0.754 $\pm$ 0.030 & 0.760 $\pm$ 0.018 \\
\midrule
\emph{SST-2, OOD inputs from AG News} & & & & \\
marginal likelihood & 0.640 $\pm$ 0.077 & 0.671 $\pm$ 0.052 & 0.766 $\pm$ 0.034 & 0.811 $\pm$ 0.006 \\
ours & \textbf{0.783} $\pm$ 0.043 & \textbf{0.768} $\pm$ 0.042 & \textbf{0.814} $\pm$ 0.024 & \textbf{0.822} $\pm$ 0.004 \\
validation search & 0.760 $\pm$ 0.065 & 0.739 $\pm$ 0.043 & 0.791 $\pm$ 0.030 & 0.812 $\pm$ 0.006 \\
\bottomrule

\end{tabular}
\end{table}

\begin{figure}[h]
\centering
\includegraphics{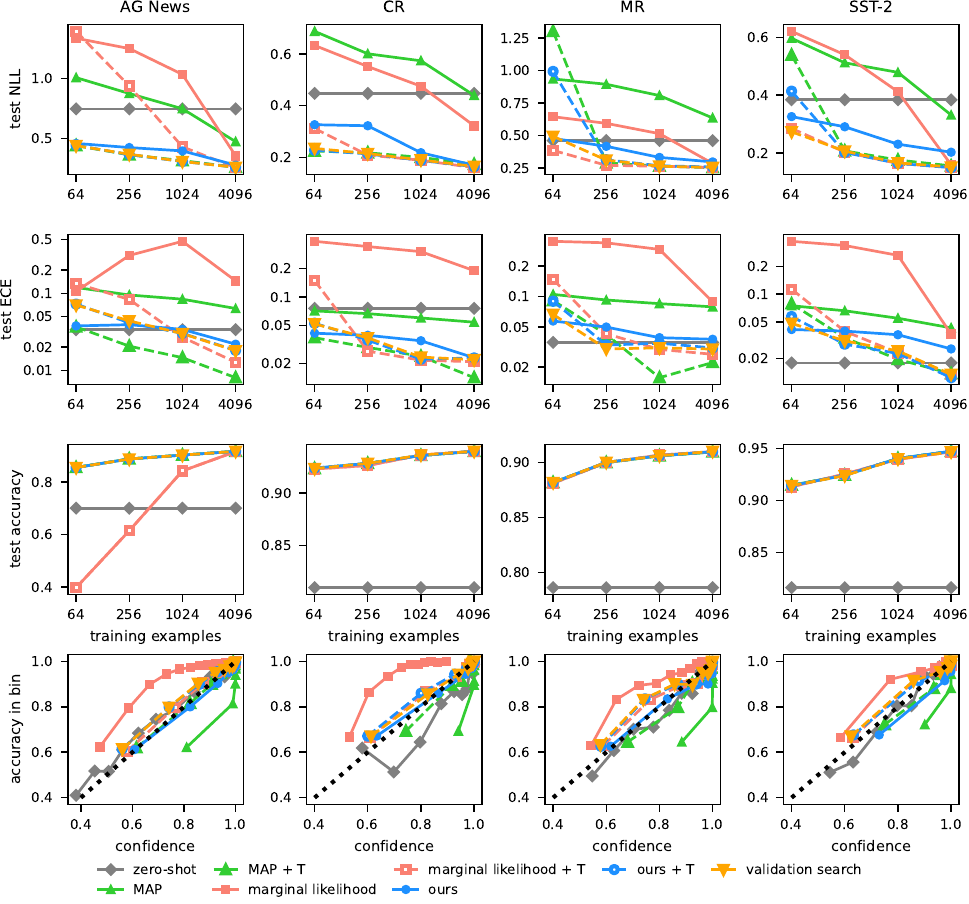}
\caption{Figure~\ref{fig:mlm} for each task, with all methods, mean over three seeds.}
\label{fig:mlm-tasks}
\end{figure}

\begin{figure}[h]
\centering
\includegraphics{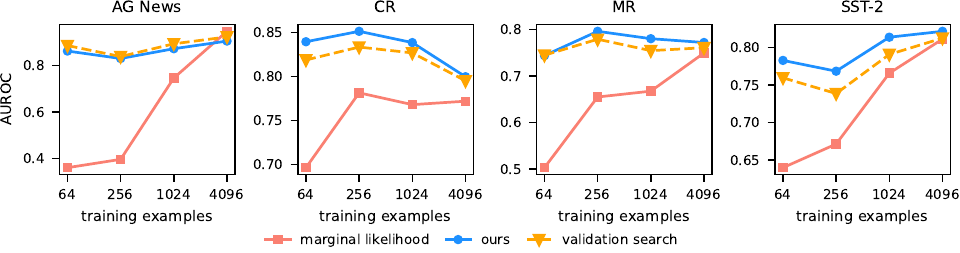}
\caption{Figure~\ref{fig:ood} for each task, mean over three seeds. The OOD inputs are SST-2 for
AG News and AG News for the other tasks.}
\label{fig:ood-tasks}
\end{figure}
\clearpage

\section{Fine-tuning under the estimated prior}
\label{app:train-lambda}

Since $\lhat$ is available before fine-tuning, it can also be used as the prior during training
(Section~\ref{sec:implementation}), not only in the Laplace approximation fitted after
training (Section~\ref{sec:exp-mlm}). The training then minimizes the sum of the training NLLs plus
$\tfrac{\lhat}{2}\lVert w\rVert^2$, the negative log density of the prior
$\mathcal N(0, \lhat^{-1} I)$ on the LoRA parameters $w$, so that the trained model is the mode of the
posterior under the estimated prior. All other settings follow Section~\ref{sec:exp-mlm} and Appendix~\ref{app:mlm}.

\begin{figure}[h]
\centering
\includegraphics{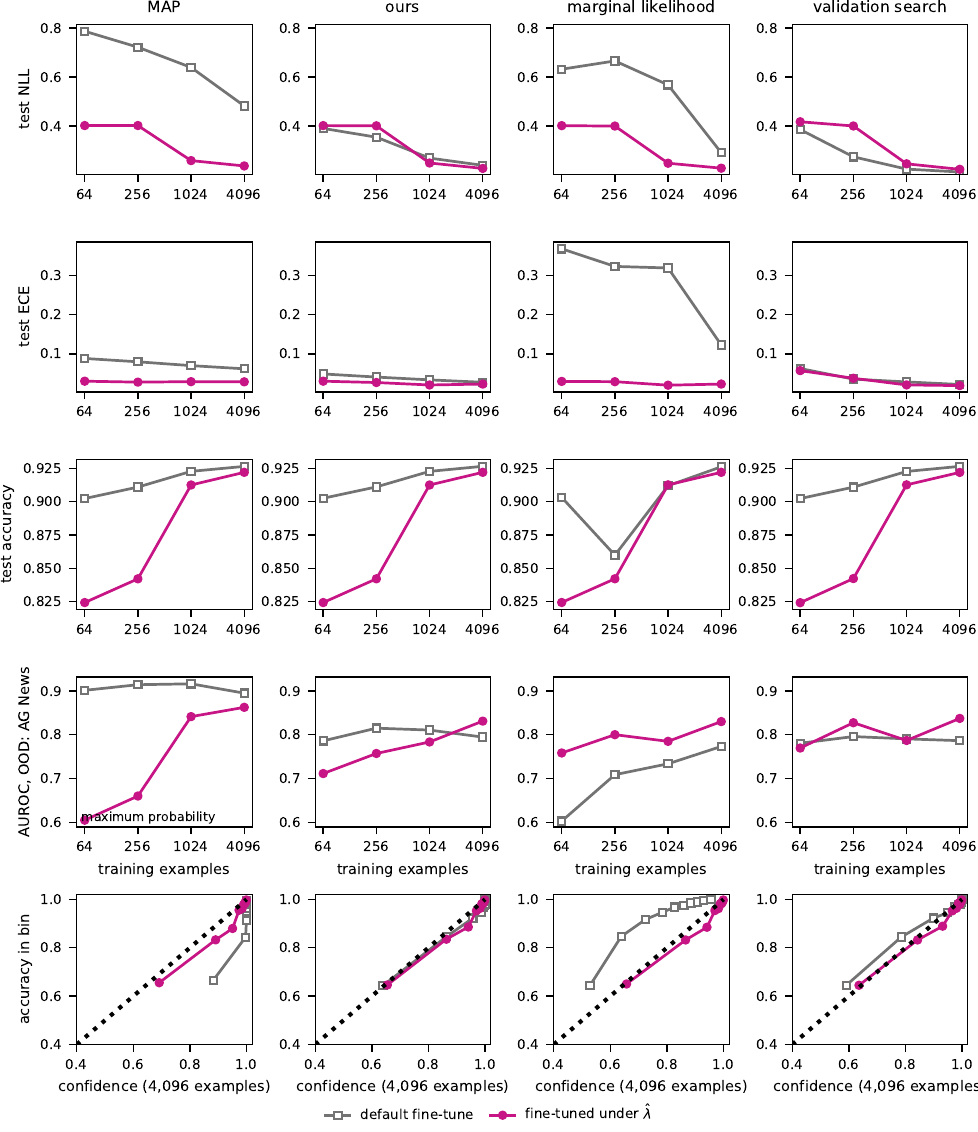}
\caption{Each method with LoRA trained without a prior (gray) and under $\lhat$ (magenta),
mean over tasks and seeds. The OOD AUROC of the MAP, which has no posterior, uses the maximum probability.
Bottom row: reliability diagrams at $4{,}096$ training examples.}
\label{fig:train-lambda}
\end{figure}

Figure~\ref{fig:train-lambda} compares each method with LoRA trained without and under the prior. The
prior improves calibration. In the reliability diagrams (bottom row), the overconfident MAP and the
underconfident marginal likelihood both move to the diagonal, and the ECE of both drops sharply. With the MAP itself well calibrated, the precision chosen after training matters
much less, and all methods reach a similar NLL. Our method maintains its NLL and improves its calibration
slightly, while validation search, which no longer has to temper an overconfident MAP, does not
benefit. Accuracy is lower for the smallest training sets, where the prior keeps the adapters close to
the pretrained model, and changes little from about a thousand training examples on.

The largest change is for marginal likelihood. With LoRA trained without a prior, it selects precisions orders of
magnitude below $\lhat$. With LoRA trained under $\lhat$, it
selects precisions close to $\lhat$ and matches our method in NLL and calibration, and with little
data it detects OOD inputs better than our method with the same training. The Laplace evidence assumes
that the trained model is the mode of the posterior under the prior it evaluates
(Appendix~\ref{app:kfac}), which holds only when LoRA is trained under a prior. Therefore, the two methods can be 
combined to improve the overall behaviour: $\lhat$ sets the prior before and during training, and marginal likelihood refines
the precision afterwards, without validation data.

\end{document}